\documentclass[11pt,oneside]{article} 
\usepackage[utf8]{inputenc} 

\usepackage{graphicx}

\usepackage[
  hmarginratio={1:1},     
  vmarginratio={1:1},     
  textwidth=400pt,        
  heightrounded,          
]{geometry}
\usepackage{amsmath}
\usepackage{booktabs} 
\usepackage{array} 
\usepackage{paralist} 
\usepackage{verbatim} 
\usepackage{subfig} 
\usepackage{setspace}
\usepackage[showzone=false,english,useregional]{datetime2}
\usepackage{tikz}
\usepackage{caption}
\usetikzlibrary{matrix,positioning,arrows.meta}
\usetikzlibrary{arrows}
\title{A One-to-One Correspondence between Natural Numbers and Binary Trees}

\begin{document}
\author{Osvaldo Skliar\thanks{Universidad Nacional, Heredia, Costa Rica. E-mail: osvaldoskliar@gmail.com. ORCID: 0000-0002-8321-3858.} \and
 Sherry Gapper \thanks{Universidad Nacional, Heredia, Costa Rica. E-mail: sherry.gapper.morrow@una.ac.cr. ORCID: 0000-0003-4920-6977.}
\and Ricardo E. Monge \thanks{Universidad Cenfotec, San Jos\'e, Costa Rica. E-mail: mongegapper@gmail.com. ORCID: 0000-0002-4321-5410.}}
\date{\today}
\maketitle

\begin{abstract}
A characterization is provided for each natural number except one (1) by means of an ordered pair of elements. The first element is a natural number called the \textit{type} of the natural number characterized, and the second is a natural number called the \textit{order} of the number characterized within those of its \textit{type}. A one-to-one correspondence is specified between the set of binary trees such that a) a given node has no child nodes (that is, it is a terminal node), or b) it has exactly two child nodes. Thus, binary trees such that one of their parent nodes has only one child node are excluded from the set considered here.
\end{abstract}



\section{Introduction}

Regarding the characterization and concept of binary tree, consult \cite[p. 363]{tree1}, \cite[p. 620]{tree2}, \cite[p. 246]{tree3} and \cite[p. 352]{tree4}.

About the use of binary trees in computer science and in artificial intelligence, consult \cite[p. 167]{app1}, \cite[p. 426]{app2}, \cite{app3}, \cite{app4}, \cite{app5}, \cite[ch. 14.4]{app6} and \cite{app7}.

A number of articles have been published on diverse correspondences and other relations between trees and natural numbers, such as \cite{prev1}, \cite{prev2}, \cite{prev3}, \cite{prev4}, \cite{prev5}, \cite{prev6} and \cite{prev7}, but these studies do not focus specifically on binary trees.

The comprehension of this article requires only a knowledge of basic notions of mathematical entities and data structures called \textit{binary trees}, including ``root node'', ``parent node'' and ``child node''.

In this article consideration is given only to binary trees with nodes fulfilling one of the following conditions: a) that the given node has no ``child nodes'', in which case it would be known as a \textit{terminal node} (or \textit{leaf node}); or b) that the given node has exactly two descendants, and would be known as a \textit{parent node}. Thus, no node in the binary trees considered can be an only child node.

The objective of this article is to establish a one-to-one (biunivocal) correspondence between binary trees and natural numbers. A unique binary tree will correspond to each natural number -- $1,2,3,\ldots$ -- and a unique natural number will correspond to any binary tree previously specified. Two variants of this one-to-one correspondence will also be described below.

To establish this one-to-one correspondence, a procedure is described in section 2 to assign an ordered pair to each natural number (other than the natural number 1). The first element of the ordered pair will be characterized as the \textit{type} of the natural number, and the second element of the given ordered pair will be called the \textit{order} of that number within those of the same \textit{type}. Both type and order values are also natural numbers.

The special case of natural number 1, within this approach, will also be addressed in section 2.

In section 3, an algorithmic procedure is described to systematically obtain the characterization of the natural numbers discussed in section 2.

In section 4, two variants are presented for the one-to-one mapping between binary trees and natural numbers.

In section 5, two possible research topics are outlined for future papers, given the results obtained so far.

\section{Type of Each Natural Number and Order of Each Number within the Respective Type}

\subsection{Type Corresponding to Each Natural Number}

A natural number, other than 1, is considered prime if it is divisible by a) itself, and b) the natural number 1.

Accept that each prime number is type 1. Thus, for example, each of the prime numbers $2, 3, 5, 7, 11$ and $13$ are considered type 1; and accept that any other prime number is also type 1. Any number that is not prime is not type 1.

Consider the number $8$. If it is expressed as a product of its prime factors, the following result is obtained:
\begin{equation*}
8=2\cdot 2 \cdot 2 = 2^{3}
\end{equation*}

Consider the number $81$. If it is expressed as a product of its prime factors, the following result is obtained:
\begin{equation*}
81=3\cdot 3 \cdot 3\cdot 3 = 3^{4}
\end{equation*}

Consider the number $49$. If it is expressed as a product of its prime factors, the following result is obtained:
\begin{equation*}
49=7\cdot 7 = 7^{2}
\end{equation*}

In these three cases, those of the natural numbers $8, 81$ and $49$, it can be observed that when expressing the product of its prime factors, the result is a single prime number raised to a power equal to or greater than $2$. 

Each of these three numbers are classified as type 2. In general, any natural number that can be expressed as a single prime factor raised to a power $n$ equal to or greater than 2 ($n=2, 3, 4, \ldots$) is considered type 2.

Consider the number $6$. When it is expressed as a product of its prime factors, the following result is obtained:
\begin{equation*}
6=2\cdot 3
\end{equation*}

Consider the number $36$. If it is expressed as a product of its prime factors, the following result is obtained:
\begin{equation*}
36=2\cdot 2 \cdot 3 \cdot 3= 2^2\cdot 3^2
\end{equation*}

Consider the number $52$. If it is expressed as a product of its prime factors, the following result is obtained:
\begin{equation*}
52=2\cdot 2\cdot 13= 2^2 \cdot 13
\end{equation*}

Each of the numbers 6, 36 and 52 is classified as type 3. In general, if a natural number expressed as a product of its prime factors has two prime factors, each of which has a power equal to or greater than one, that natural number is type 3.

The rule used to determine the type corresponding to any natural number equal to $2$ or greater than $2$ ($n\geq 2$) is the following:
\begin{enumerate}
\item Any prime number is considered type 1; and
\item The type of any non-prime natural number $n$ greater than 2 ($n > 2$) is equal to the number of prime factors of which it is a multiple plus 1. 
\end{enumerate}

Consider the natural number $729$. It may be expressed as follows:
\begin{equation*}
729= 3^6
\end{equation*} 

This number, $729$, is a multiple of a single prime: the number 3. Of course, $729$ is also a multiple of other numbers (such as 9 and 81), but these numbers are not primes. When the above rule is applied to the number $729$ to determine its type, it is type 2. That type is 1 more than the number of unique prime numbers (also 1) of which it is a multiple.

Consider the number $490$. It may be expressed as follows:
\begin{equation*}
490= 2\cdot 5 \cdot 7^2
\end{equation*}

Since $490$ is a multiple of three prime numbers (2, 5 and 7), its type is $3+1$, or 4, according to the respective rule.

\subsection{Order Corresponding to Each Natural Number $n$ for $n\geq 2$, within the Respective Type}

Consider the set of prime numbers: $2,3,5,\ldots\;$. As accepted above, each of these numbers is type 1. The number $2$ is first; so its order, within those in type 1, is $1$. The number $3$ is the second of those in type 1; thus, its order will be 2. The number $5$ is the third type 1 number; thus, its order number will be 3. The number $7$ is the fourth type 1 number; thus, its order number will be 4; and so on.

A similar approach is used to assign an order number to numbers whose type is greater than 1.

Consider number 78. This number may be expressed as follows:
\begin{equation*}
78= 2\cdot 3 \cdot 13
\end{equation*}

All three numbers (2, 3 and 13) are primes. Using the rule provided, the type of $78$ is one more than 3; that is, 4. Within type 4 natural numbers, 78 is the sixth one. Thus, its order number is $6$. Five type 4 numbers are less than $78$: 30, since $30=2\cdot 3\cdot 5$; 42, since $42=2\cdot 3\cdot 7$; 60, since $60=2^2\cdot 3\cdot 5$; 66, since $66=2\cdot 3\cdot 11$ and 70, since $70=2\cdot 5\cdot 7$. Therefore, the number $78$ is assigned order number 6, within the numbers of its type.

As indicated in section 1, section 3 introduces an algorithmic procedure to characterize each natural number $n$, for $n\geq 2$, with an ordered pair. The first element of the ordered pair is the \textit{type} corresponding to that natural number and the second element of the ordered pair is the \textit{order number} corresponding to that natural number, within its \textit{type}.

This form of characterization for any natural number $n$, such that $n\geq 2$, is exemplified as follows: The natural number $117$ can be characterized by the ordered pair $(3,64)$. The natural number $598$ can be characterized by the ordered pair $(4,143)$. The natural number $798$ can be characterized by the ordered pair $(5,15)$.

Specific attention should be given to the natural number 1, according to the approach presented in this article. The natural number 1, shares a feature with prime numbers: It is divisible by itself and by $1$. Nevertheless, it is different from the primes because for 1 both numbers are the same.

It was seen above that every prime number is type 1. The number following 1 in the natural numbers is $2$; it is also a prime, and hence type 1. Its order number, within those in type 1, is 1. That is according to the criterion described above, the characterization of the number 2 is $(1,1)$.

The number 1 precedes the number 2, and it is not type 1 because it is not prime; nor is it a multiple of another prime. According to the rule established above to determine the type of each natural number, neither a type nor an order number may be assigned to the number 1. It would be senseless to assign an order number to an inexistent type. This, however, according to the criteria presented in this article, does not affect the possibility of establishing a one-to-one correspondence between natural numbers and binary trees, as described in section 4. 

\section{An Algorithmic Procedure to Characterize the Natural Numbers (Other than 1)}

To explain this algorithmic approach, it is applied to the first 30 natural numbers: $1,2,3,\ldots,30$.

Suppose that each of these first 30 natural numbers, placed one after another in ascending order from 1 to 30, occupies a spatial region limited by side ``walls''. Each space will be called a \textit{cell}. Thus, there will be a cell for 1, a cell for 2, another for 3, and so on successively.

If this algorithmic approach is applied to a different number of cells, the process must include a number of cells equal to that number. For example, to illustrate the algorithmic approach for the number $1,573$, there must be $1,573$ cells. Figure \ref{f1} displays the cells for $n=30$.
 
\begin{figure}[!h] 
\centering
\begin{tikzpicture}
\matrix[inner sep=0pt] (S) 
       [matrix of math nodes,nodes in empty cells,
       nodes={outer sep=0pt,minimum width=10mm,minimum height=7mm}]
      { 1  & 2  & 3 & 4 & 5 & 6 & 7 & 8 & 9 & 10  \\ 
			     &    &    &   &   &  &   &    &   &    \\ 
           &    &    &   &   &  &   &   &   &   \\ 
           &    &    &   &   &  &   &   &   &   \\  
			     &    &    &   &   &  &   &    &   &    \\ 
           &    &    &   &   &  &   &   &   &   \\ };
\foreach \i in {1,...,10} {
		\draw[solid,black] (S-1-\i.north west) -- (S-6-\i.north west);
		\draw[dashed,black] (S-6-\i.north west) -- (S-6-\i.south west);
}

		\draw[solid,black] (S-1-10.north east) -- (S-6-10.north east);
		\draw[dashed,black] (S-6-10.north east) -- (S-6-10.south east);

\matrix[below=1cm of S,inner sep=0pt] (SS)  
       [matrix of math nodes,nodes in empty cells,
       nodes={outer sep=0pt,minimum width=10mm,minimum height=7mm}]
      { 11  & 12  & 13 & 14 & 15 & 16 & 17 & 18 & 19 & 20  \\ 
			     &    &    &   &   &  &   &    &   &    \\ 
           &    &    &   &   &  &   &   &   &   \\ 
           &    &    &   &   &  &   &   &   &   \\  
			     &    &    &   &   &  &   &    &   &    \\ 
           &    &    &   &   &  &   &   &   &   \\ };
\foreach \i in {1,...,10} {
		\draw[solid,black] (SS-1-\i.north west) -- (SS-6-\i.north west);
		\draw[dashed,black] (SS-6-\i.north west) -- (SS-6-\i.south west);
}

		\draw[solid,black] (SS-1-10.north east) -- (SS-6-10.north east);
		\draw[dashed,black] (SS-6-10.north east) -- (SS-6-10.south east);

\matrix[below=1cm of SS,inner sep=0pt] (SSS)  
       [matrix of math nodes,nodes in empty cells,
       nodes={outer sep=0pt,minimum width=10mm,minimum height=7mm}]
      { 21  & 22  & 23 & 24 & 25 & 26 & 27 & 28 & 29 & 30  \\ 
			     &    &    &   &   &  &   &    &   &    \\ 
           &    &    &   &   &  &   &   &   &   \\ 
           &    &    &   &   &  &   &   &   &   \\  
			     &    &    &   &   &  &   &    &   &    \\ 
           &    &    &   &   &  &   &   &   &   \\ };
\foreach \i in {1,...,10} {
		\draw[solid,black] (SSS-1-\i.north west) -- (SSS-6-\i.north west);
		\draw[dashed,black] (SSS-6-\i.north west) -- (SSS-6-\i.south west);
}

		\draw[solid,black] (SSS-1-10.north east) -- (SSS-6-10.north east);
		\draw[dashed,black] (SSS-6-10.north east) -- (SSS-6-10.south east);

\end{tikzpicture}
\caption{\textit{Cells} corresponding to each of the natural numbers considered from 1 to 30} 
\label{f1}
\end{figure}
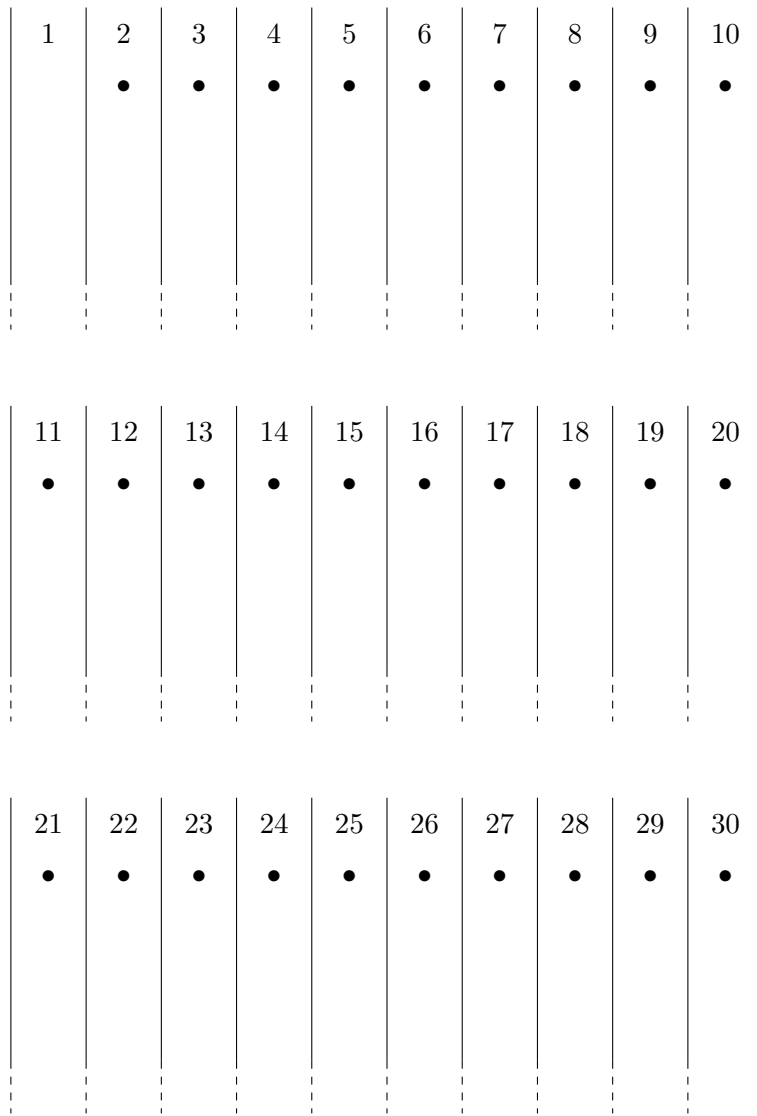
\begin{figure}[!h] 
\centering
\begin{tikzpicture}
\matrix[inner sep=0pt] (S) 
       [matrix of math nodes,nodes in empty cells,
       nodes={outer sep=0pt,minimum width=10mm,minimum height=7mm}]
      { 1  & 2  & 3 & 4 & 5 & 6 & 7 & 8 & 9 & 10  \\ 
 & \bullet & \bullet & \bullet & \bullet & \bullet & \bullet & \bullet & \bullet & \bullet \\
           &    &    &   &   &  &   &   &   &   \\ 
           &    &    &   &   &  &   &   &   &   \\  
			     &    &    &   &   &  &   &    &   &    \\ 
           &    &    &   &   &  &   &   &   &   \\ };
\foreach \i in {1,...,10} {
		\draw[solid,black] (S-1-\i.north west) -- (S-6-\i.north west);
		\draw[dashed,black] (S-6-\i.north west) -- (S-6-\i.south west);
}

		\draw[solid,black] (S-1-10.north east) -- (S-6-10.north east);
		\draw[dashed,black] (S-6-10.north east) -- (S-6-10.south east);

\matrix[below=1cm of S,inner sep=0pt] (SS)  
       [matrix of math nodes,nodes in empty cells,
       nodes={outer sep=0pt,minimum width=10mm,minimum height=7mm}]
      { 11  & 12  & 13 & 14 & 15 & 16 & 17 & 18 & 19 & 20  \\ 
			   \bullet & \bullet & \bullet & \bullet & \bullet & \bullet & \bullet & \bullet & \bullet & \bullet   \\ 
           &    &    &   &   &  &   &   &   &   \\ 
           &    &    &   &   &  &   &   &   &   \\  
			     &    &    &   &   &  &   &    &   &    \\ 
           &    &    &   &   &  &   &   &   &   \\ };
\foreach \i in {1,...,10} {
		\draw[solid,black] (SS-1-\i.north west) -- (SS-6-\i.north west);
		\draw[dashed,black] (SS-6-\i.north west) -- (SS-6-\i.south west);
}

		\draw[solid,black] (SS-1-10.north east) -- (SS-6-10.north east);
		\draw[dashed,black] (SS-6-10.north east) -- (SS-6-10.south east);

\matrix[below=1cm of SS,inner sep=0pt] (SSS)  
       [matrix of math nodes,nodes in empty cells,
       nodes={outer sep=0pt,minimum width=10mm,minimum height=7mm}]
      { 21  & 22  & 23 & 24 & 25 & 26 & 27 & 28 & 29 & 30  \\ 
			   \bullet & \bullet & \bullet & \bullet & \bullet & \bullet & \bullet & \bullet & \bullet & \bullet \\
   &    &    &   &   &  &   &   &   &   \\ 
           &    &    &   &   &  &   &   &   &   \\  
			     &    &    &   &   &  &   &    &   &    \\ 
           &    &    &   &   &  &   &   &   &   \\ };
\foreach \i in {1,...,10} {
		\draw[solid,black] (SSS-1-\i.north west) -- (SSS-6-\i.north west);
		\draw[dashed,black] (SSS-6-\i.north west) -- (SSS-6-\i.south west);
}

		\draw[solid,black] (SSS-1-10.north east) -- (SSS-6-10.north east);
		\draw[dashed,black] (SSS-6-10.north east) -- (SSS-6-10.south east);

\end{tikzpicture}
\caption{Here a dot has been added to each of the cells shown in figure \ref{f1}, corresponding to multiples of $1$; that is, to the numbers $2,3,4,\ldots,30$.} 
\label{f2}
\end{figure}

\newpage 
Figure \ref{f1} is reproduced in figure \ref{f2} with a slight modification: A dot has been added to each of the cells shown in figure \ref{f1}, to represent the multiples of $1$ graphically. In this case, these multiples are the numbers $2,3,4,\ldots,30$.
\newpage 

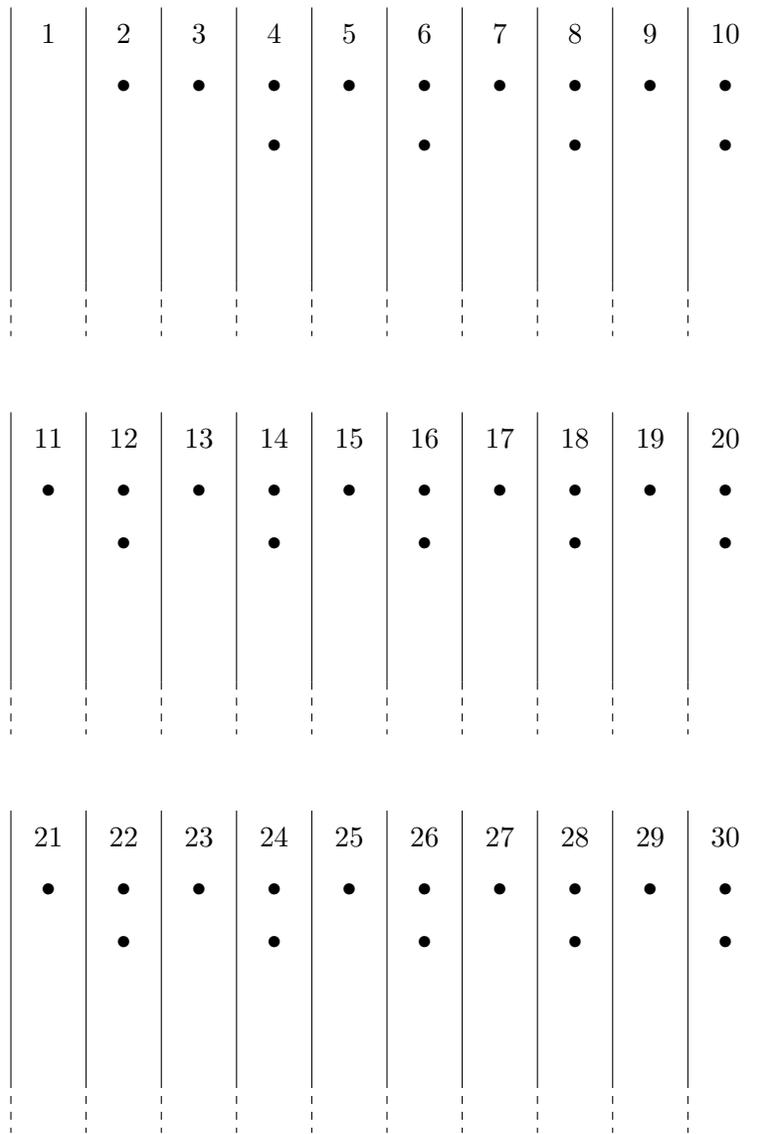
\begin{figure}[!h] 
\centering
\begin{tikzpicture}
\matrix[inner sep=0pt] (S) 
       [matrix of math nodes,nodes in empty cells,
       nodes={outer sep=0pt,minimum width=10mm,minimum height=7mm}]
      { 1  & 2  & 3 & 4 & 5 & 6 & 7 & 8 & 9 & 10  \\ 
 & \bullet & \bullet & \bullet & \bullet & \bullet & \bullet & \bullet & \bullet & \bullet \\
           &    &    & \bullet  &   & \bullet &   & \bullet  &   & \bullet  \\ 
           &    &    &   &   &  &   &   &   &   \\  
			     &    &    &   &   &  &   &    &   &    \\ 
           &    &    &   &   &  &   &   &   &   \\ };
\foreach \i in {1,...,10} {
		\draw[solid,black] (S-1-\i.north west) -- (S-6-\i.north west);
		\draw[dashed,black] (S-6-\i.north west) -- (S-6-\i.south west);
}

		\draw[solid,black] (S-1-10.north east) -- (S-6-10.north east);
		\draw[dashed,black] (S-6-10.north east) -- (S-6-10.south east);

\matrix[below=1cm of S,inner sep=0pt] (SS)  
       [matrix of math nodes,nodes in empty cells,
       nodes={outer sep=0pt,minimum width=10mm,minimum height=7mm}]
      { 11  & 12  & 13 & 14 & 15 & 16 & 17 & 18 & 19 & 20  \\ 
			   \bullet & \bullet & \bullet & \bullet & \bullet & \bullet & \bullet & \bullet & \bullet & \bullet   \\ 
           & \bullet   &    & \bullet  &   & \bullet &   &\bullet   &   & \bullet  \\ 
           &    &    &   &   &  &   &   &   &   \\  
			     &    &    &   &   &  &   &    &   &    \\ 
           &    &    &   &   &  &   &   &   &   \\ };
\foreach \i in {1,...,10} {
		\draw[solid,black] (SS-1-\i.north west) -- (SS-6-\i.north west);
		\draw[dashed,black] (SS-6-\i.north west) -- (SS-6-\i.south west);
}

		\draw[solid,black] (SS-1-10.north east) -- (SS-6-10.north east);
		\draw[dashed,black] (SS-6-10.north east) -- (SS-6-10.south east);

\matrix[below=1cm of SS,inner sep=0pt] (SSS)  
       [matrix of math nodes,nodes in empty cells,
       nodes={outer sep=0pt,minimum width=10mm,minimum height=7mm}]
      { 21  & 22  & 23 & 24 & 25 & 26 & 27 & 28 & 29 & 30  \\ 
			   \bullet & \bullet & \bullet & \bullet & \bullet & \bullet & \bullet & \bullet & \bullet & \bullet \\
   &  \bullet  &    & \bullet  &   & \bullet &   & \bullet  &   &\bullet   \\ 
           &    &    &   &   &  &   &   &   &   \\  
			     &    &    &   &   &  &   &    &   &    \\ 
           &    &    &   &   &  &   &   &   &   \\ };
\foreach \i in {1,...,10} {
		\draw[solid,black] (SSS-1-\i.north west) -- (SSS-6-\i.north west);
		\draw[dashed,black] (SSS-6-\i.north west) -- (SSS-6-\i.south west);
}

		\draw[solid,black] (SSS-1-10.north east) -- (SSS-6-10.north east);
		\draw[dashed,black] (SSS-6-10.north east) -- (SSS-6-10.south east);

\end{tikzpicture}

\caption{This figure is based on figure \ref{f2}, with a dot added to each multiple of $2$: $4,6,8,\ldots,30$.} 
\label{f3}
\end{figure}

Consider the number following 1 (i.e., 2). Is there only one dot in the cell corresponding to $2$? If the answer were negative, one would continue with the number following $2$ ($3$), without adding anything to figure \ref{f2}. However, since there is one dot in the cell for $2$, the answer is affirmative. A dot is added to each cell corresponding to a multiple of 2, from 3 to 30 inclusive. This is illustrated in figure \ref{f3}.

\newpage 

Consider the next cell after $2$: that of the number 3. Is there only one dot in the cell corresponding to $3$? If the answer were negative, one would continue with the number following 3 ($4$), without adding anything to figure \ref{f3}. However, the answer is affirmative, because there is only one dot in the cell for $3$. A dot is added to each of the cells of numbers corresponding to a multiple of 3, from 4 to 30 inclusive. These cells correspond to the numbers $6, 9, 12, 15, \ldots,30$.

Continue using the same procedure, examining one cell after another until the cell corresponding to $\tfrac{30}{2}$; that is, up to 15 inclusive. If there is no dot, nothing is added to the last figure. If, however, in any of the cells examined corresponding to a number $n$, such that $n$ is less than or equal to 15 ($n\leq 15$), there is a dot, then a dot must be added to the set of all cells corresponding to numbers that are multiples of $n$, up to 30 inclusive.

Figure \ref{f4} represents the results of running the entire procedure for the cells specified.

\begin{figure}[!h] 
\centering
\begin{tikzpicture}
\matrix[inner sep=0pt] (S) 
      [matrix of math nodes,nodes in empty cells,
       nodes={outer sep=0pt,minimum width=10mm,minimum height=7mm}]
      { 1  & 2  & 3 & 4 & 5 & 6 & 7 & 8 & 9 & 10  \\ 
  & \bullet & \bullet & \bullet & \bullet & \bullet & \bullet & \bullet & \bullet & \bullet \\
           &    &    & \bullet  &   & \bullet &   & \bullet  & \bullet  & \bullet  \\ 
           &    &    &   &   &\bullet  &   &   &   &  \bullet  \\  
			     &    &    &   &   &  &   &    &   &    \\ 
           &    &    &   &   &  &   &   &   &   \\ };
\foreach \i in {1,...,10} {
		\draw[solid,black] (S-1-\i.north west) -- (S-6-\i.north west);
		\draw[dashed,black] (S-6-\i.north west) -- (S-6-\i.south west);
}

		\draw[solid,black] (S-1-10.north east) -- (S-6-10.north east);
		\draw[dashed,black] (S-6-10.north east) -- (S-6-10.south east);

\matrix[below=1cm of S,inner sep=0pt] (SS)  
       [matrix of math nodes,nodes in empty cells,
       nodes={outer sep=0pt,minimum width=10mm,minimum height=7mm}]
      { 11  & 12  & 13 & 14 & 15 & 16 & 17 & 18 & 19 & 20  \\ 
			   \bullet & \bullet & \bullet & \bullet & \bullet & \bullet & \bullet & \bullet & \bullet & \bullet   \\ 
           & \bullet   &    & \bullet  & \bullet   & \bullet &   &\bullet   &   & \bullet  \\ 
           & \bullet   &    & \bullet  & \bullet  &  &   & \bullet  &   & \bullet  \\  
			     &    &    &   &   &  &   &    &   &    \\ 
           &    &    &   &   &  &   &   &   &   \\ };
\foreach \i in {1,...,10} {
		\draw[solid,black] (SS-1-\i.north west) -- (SS-6-\i.north west);
		\draw[dashed,black] (SS-6-\i.north west) -- (SS-6-\i.south west);
}

		\draw[solid,black] (SS-1-10.north east) -- (SS-6-10.north east);
		\draw[dashed,black] (SS-6-10.north east) -- (SS-6-10.south east);

\matrix[below=1cm of SS,inner sep=0pt] (SSS)  
       [matrix of math nodes,nodes in empty cells,
       nodes={outer sep=0pt,minimum width=10mm,minimum height=7mm}]
      { 21  & 22  & 23 & 24 & 25 & 26 & 27 & 28 & 29 & 30  \\ 
			   \bullet & \bullet & \bullet & \bullet & \bullet & \bullet & \bullet & \bullet & \bullet & \bullet \\
  \bullet &  \bullet  &    & \bullet  &  \bullet & \bullet & \bullet  & \bullet  &   &\bullet   \\ 
       \bullet    & \bullet   &    & \bullet  &   & \bullet &   &  \bullet &   & \bullet  \\  
			     &    &    &   &   &  &   &    &   &   \bullet \\ 
           &    &    &   &   &  &   &   &   &   \\ };
\foreach \i in {1,...,10} {
		\draw[solid,black] (SSS-1-\i.north west) -- (SSS-6-\i.north west);
		\draw[dashed,black] (SSS-6-\i.north west) -- (SSS-6-\i.south west);
}

		\draw[solid,black] (SSS-1-10.north east) -- (SSS-6-10.north east);
		\draw[dashed,black] (SSS-6-10.north east) -- (SSS-6-10.south east);

\end{tikzpicture}
\caption{The results of the entire procedure are shown here.} 
\label{f4}
\end{figure}
\newpage 

The \textit{type} of each of the natural numbers $2,3,4,\ldots,30$ is equal to the number of dots in their respective cells. For example, there are three dots in the cell corresponding to the number 18; thus the type for number 18 is 3. There are four dots in the cell corresponding to number 30; thus the type for the number 30 is 4.

The \textit{order} of each number ($2,3,4,\ldots,30$) is determined as specified above. For example, the order of the number 18 (within the type 3 numbers) is $6$ because it is preceded by five other type 3 numbers: $6,10,12,14$ and $15$. The order of the number 30 (within type 4 numbers) is $1$ because it is the first type 4 number among the 30 numbers considered.

Figure \ref{f5} reproduces figure \ref{f4} and adds, at the bottom of every cell, the ordered pair for each number, specifying its type and order.

\begin{figure}[!h] 
\centering
\begin{tikzpicture}
\matrix[inner sep=0pt] (S) 
       [matrix of math nodes,nodes in empty cells,
       nodes={outer sep=0pt,minimum width=10mm,minimum height=7mm}]
      { 1  & 2  & 3 & 4 & 5 & 6 & 7 & 8 & 9 & 10  \\ 
  & \bullet & \bullet & \bullet & \bullet & \bullet & \bullet & \bullet & \bullet & \bullet \\
           &    &    & \bullet  &   & \bullet &   & \bullet  & \bullet  & \bullet  \\ 
           &    &    &   &   &\bullet  &   &   &   &  \bullet  \\  
			     &    &    &   &   &  &   &    &   &    \\ 
           &  (1,1)  & (1,2)   & (2,1)  & (1,3)  & (3,1) & (1,4)  & (2,2)  & (2,3)  & (3,2)  \\ };
\foreach \i in {1,...,10} {
		\draw[solid,black] (S-1-\i.north west) -- (S-6-\i.north west);
		\draw[dashed,black] (S-6-\i.north west) -- (S-6-\i.south west);
}

		\draw[solid,black] (S-1-10.north east) -- (S-6-10.north east);
		\draw[dashed,black] (S-6-10.north east) -- (S-6-10.south east);

\matrix[below=1cm of S,inner sep=0pt] (SS)  
       [matrix of math nodes,nodes in empty cells,
       nodes={outer sep=0pt,minimum width=10mm,minimum height=7mm}]
      { 11  & 12  & 13 & 14 & 15 & 16 & 17 & 18 & 19 & 20  \\ 
			   \bullet & \bullet & \bullet & \bullet & \bullet & \bullet & \bullet & \bullet & \bullet & \bullet   \\ 
           & \bullet   &    & \bullet  & \bullet   & \bullet &   &\bullet   &   & \bullet  \\ 
           & \bullet   &    & \bullet  & \bullet  &  &   & \bullet  &   & \bullet  \\  
			     &    &    &   &   &  &   &    &   &    \\ 
          (1,5) & (3,3)   & (1,6)   & (3,4)  &(3,5)   & (2,4) & (1,7)  & (3,6)  & (1,8)  & (3,7)  \\ };
\foreach \i in {1,...,10} {
		\draw[solid,black] (SS-1-\i.north west) -- (SS-6-\i.north west);
		\draw[dashed,black] (SS-6-\i.north west) -- (SS-6-\i.south west);
}

		\draw[solid,black] (SS-1-10.north east) -- (SS-6-10.north east);
		\draw[dashed,black] (SS-6-10.north east) -- (SS-6-10.south east);

\matrix[below=1cm of SS,inner sep=0pt] (SSS)  
       [matrix of math nodes,nodes in empty cells,
       nodes={outer sep=0pt,minimum width=10mm,minimum height=7mm}]
      { 21  & 22  & 23 & 24 & 25 & 26 & 27 & 28 & 29 & 30  \\ 
			   \bullet & \bullet & \bullet & \bullet & \bullet & \bullet & \bullet & \bullet & \bullet & \bullet \\
  \bullet &  \bullet  &    & \bullet  &  \bullet & \bullet & \bullet  & \bullet  &   &\bullet   \\ 
       \bullet    & \bullet   &    & \bullet  &   & \bullet &   &  \bullet &   & \bullet  \\  
			     &    &    &   &   &  &   &    &   &   \bullet \\ 
          (3,8) & (3,9)   & (1,9)   & (3,10)  & (2,5)  & (3,11) & (2,6)  & (3,12)  & (1,10)   &  (4,1) \\ };
\foreach \i in {1,...,10} {
		\draw[solid,black] (SSS-1-\i.north west) -- (SSS-6-\i.north west);
		\draw[dashed,black] (SSS-6-\i.north west) -- (SSS-6-\i.south west);
}

		\draw[solid,black] (SSS-1-10.north east) -- (SSS-6-10.north east);
		\draw[dashed,black] (SSS-6-10.north east) -- (SSS-6-10.south east);

\end{tikzpicture}
\caption{The ordered pairs indicating type and order of the natural numbers (from 2 to 30) have been added here to the cells shown in figure \ref{f4}.}
\label{f5}
\end{figure}
\newpage 

The reasoning behind the limit chosen for the first 30 natural numbers (i.e., examining only up to $\tfrac{30}{2}$, i.e. $15$) is the following: Any cell corresponding to a given number greater than 15 with only one dot generates dots for cells corresponding to multiples of that number outside the range considered. For example, if the cell corresponding to $17$ is examined, it would have only one dot, and dots would be added to the multiples of 17: $34, 51,68, \ldots$. Those numbers are not elements of the set of natural numbers considered ($1,2,3,\ldots,30$).

Suppose that instead of the first $30$ natural numbers, consideration had been given to the first $N$ natural numbers. $N$ could be any natural number. In this case, the procedure to determine whether a dot must be added should be carried out    only until the cell corresponding to the number equal to the whole part of $\tfrac{N}{2}$, or $\mathrm{Int}(\tfrac{N}{2}$). If $N$ is even, then $\mathrm{Int}(\tfrac{N}{2})$ is equal to $\tfrac{N}{2}$, or $\mathrm{Int}(\tfrac{N}{2})$=$\tfrac{N}{2}$.

Consider the first $2,548$ natural numbers. In this case, the examination process specified must continue only until the cell corresponding to the number $1,274$, since $\tfrac{2,548}{2}=1,274$.

Consider the first $35,725$ natural numbers. In this case, the examination process must continue only until the cell corresponding to the number $17, 862$, since $\mathrm{Int}(\tfrac{35,725}{2})=17,862$.
  
A list of the ordered pairs is provided in the appendix of this article for the natural numbers up to $1080$ ($2,3,4,\ldots,1080$). As seen above, with this algorithmic approach no ordered pair (type, order) corresponds to the natural number 1.

\section{A One-to-One Correspondence between Binary Trees and Natural Numbers}
\subsection{Preliminary Considerations on the Terms Used}
As seen in section 1, consideration is given in this article only to binary trees in which each node: a) has no children (a terminal node), or b) has exactly two children; i. e.: each node corresponding to the number 1 has no child nodes, and any node with a natural number other than 1 will have exactly two child nodes.

Clearly, the name ``trees'' used for certain graphs originated from the similarity between their graphical representations and real trees. Thus, the graphic representation of the mathematical entity ``tree'' has the shape of an inverted tree, with the root node at the top, and the part of the mathematical structure which is similar to the branches of a real tree extending downward.
 
Each binary tree considered will have one root node. The term ``terminal node'' will correspond to any node with no children. There is only one case, that of a binary tree corresponding to the natural number 1, in which the only node of that binary tree is both a root node and a terminal node.

Each node that is not a terminal node has exactly two child nodes, and is known as the ``parent node'' of those two descendants. The relationship between the parent node and either of its child nodes is asymmetric, and it is represented by an arrow. Thus, there is one arrow pointing from each of the parent nodes toward one of its child nodes and another arrow pointing from that parent node to the other child node.

\subsection{First Variant of the One-to-One Correspondence between Binary Trees and Natural Numbers}

First, three examples will be given of that one-to-one correspondence, using the first variant.

The first example is the number 25. According to the procedure described in section 3, this number is type 2 and its order is 5 within the numbers of its type; 25 can be represented by the ordered pair $(2,5)$.

\begin{figure}[!h] 
\centering
\includegraphics{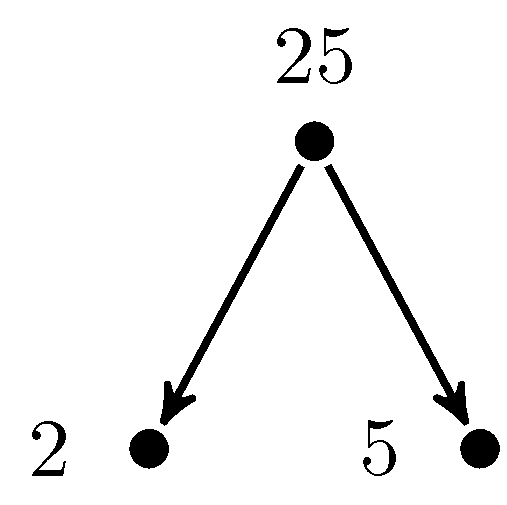}
\caption{The first part of the tree for the number 25}
\label{f6}
\end{figure}

Figure \ref{f6} displays the first part of the binary tree corresponding to number 25. Only the root node (in this case, assigned the number 25) is visible with its two children. The left child is assigned the \textit{type} of the number 25 (that is, 2) and the right child is assigned the \textit{order} of the number 25 (that is, 5). Here the root node to which the number 25 was assigned is considered the parent node of the child node to which the number 2 corresponds (the \textit{type} of the number 25), and of the child node to which the number 5 corresponds (the \textit{order} of the number 25) within those of its order.
 
Next, the two child nodes of the node that the number 2 corresponded to and the two child nodes that the number 5 corresponded to are added to the binary tree that represents $25$.

This process will be repeated for each new set of two child nodes obtained by the process described, until all are terminal nodes. Recall that the number 1 corresponds to each terminal node.

In all cases, the left child of any parent node is assigned the number corresponding to the \textit{type}, and the right child of that parent node is assigned the number corresponding to the \textit{order} of that parent node, respectively.
 
\begin{figure}[!h] 
\centering
\includegraphics{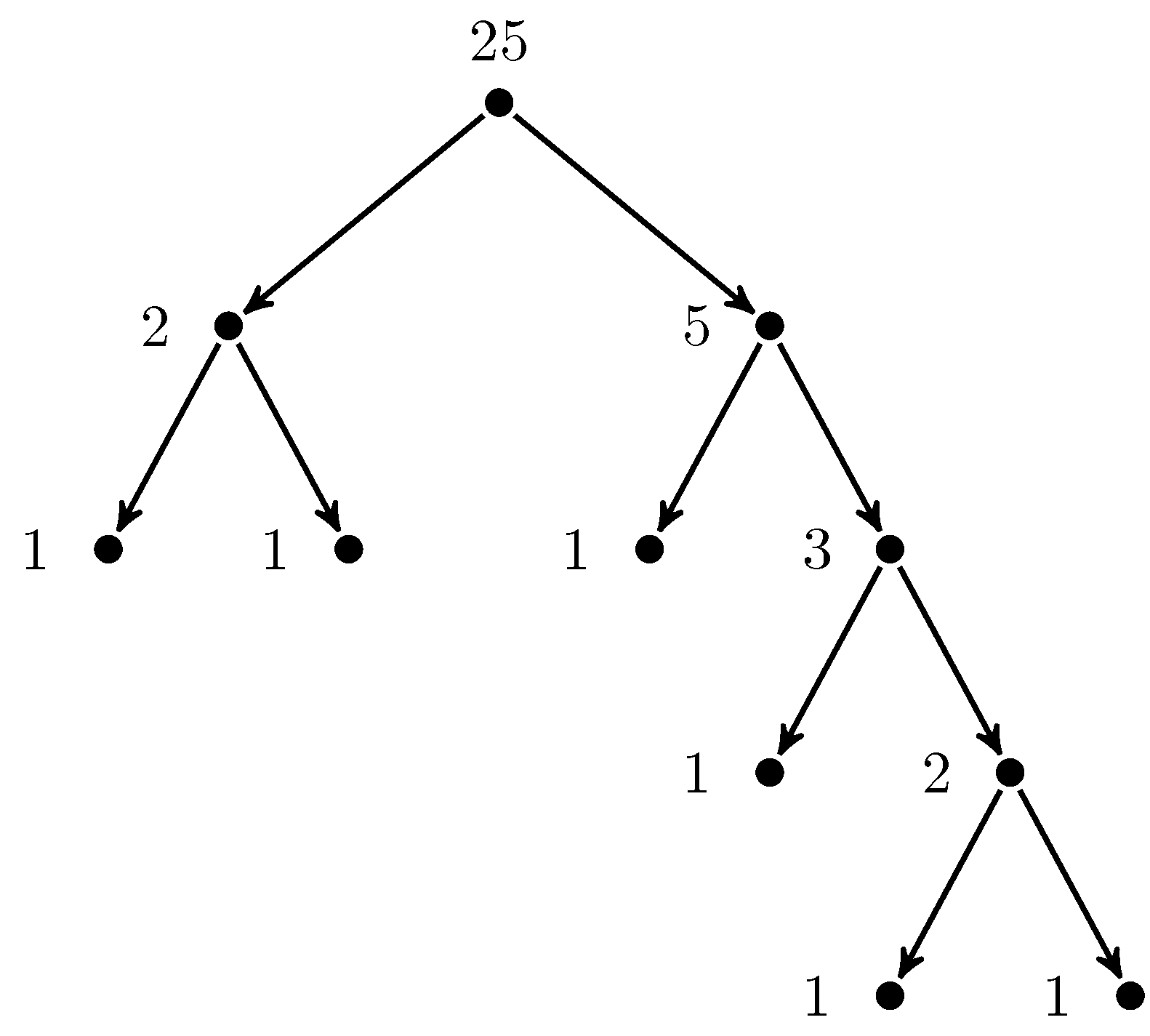}
\caption{The entire binary tree for the number 25}.
\label{f7}
\end{figure}

The complete binary tree for the number 25 is shown in figure \ref{f7}.
\newpage

\begin{figure}[!h] 
\centering
\includegraphics{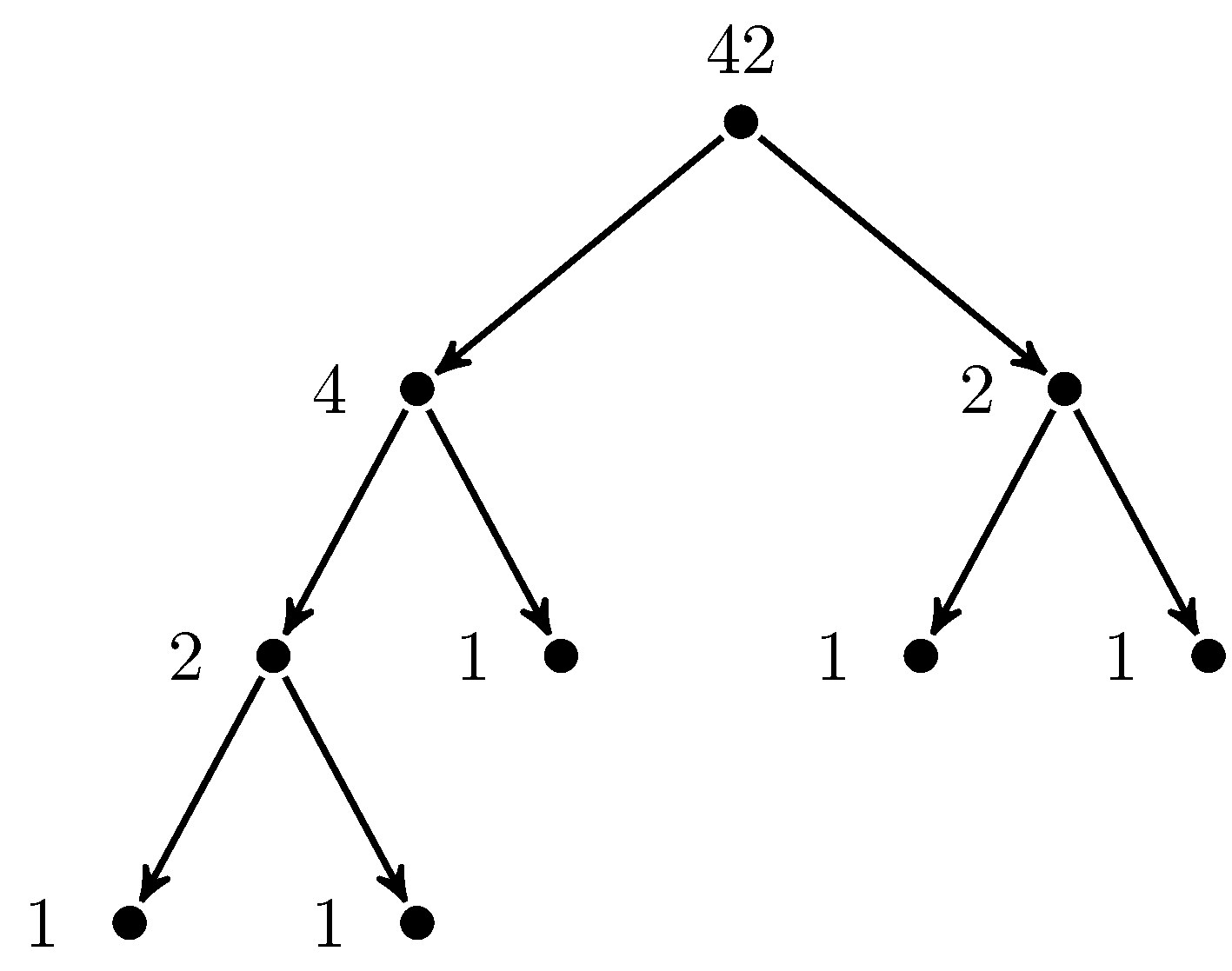}
\caption{The entire binary tree for the number 42}
\label{f8}
\end{figure}

\begin{figure}[!h] 
\centering
\includegraphics{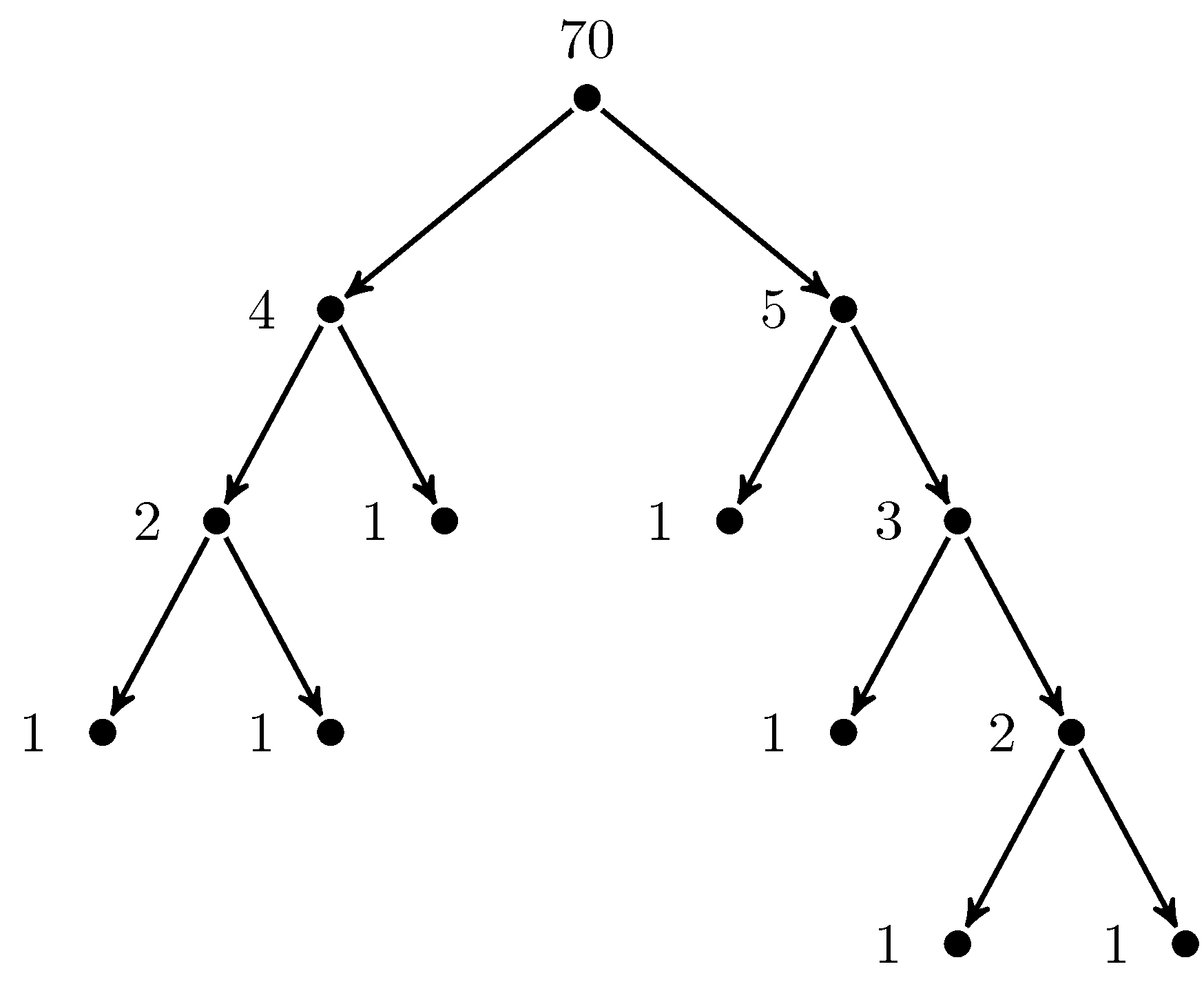}
\caption{The entire binary tree for the number 70}
\label{f9}
\end{figure}

The binary trees corresponding to the numbers 42 and 70, constructed using the same procedure, have been represented in figures \ref{f8} and \ref{f9}.

\newpage
Siince the natural number 1 is assigned to each terminal node of a binary tree with the characteristics described in section 1, it is possible to determine the natural number for each terminal node and begin labelling the parent nodes starting with the terminal nodes. If the numbers corresponding to any pair of child nodes (siblings) are known, the number corresponding to their parent node can be determined.

\begin{figure}[!h] 
\centering
\includegraphics{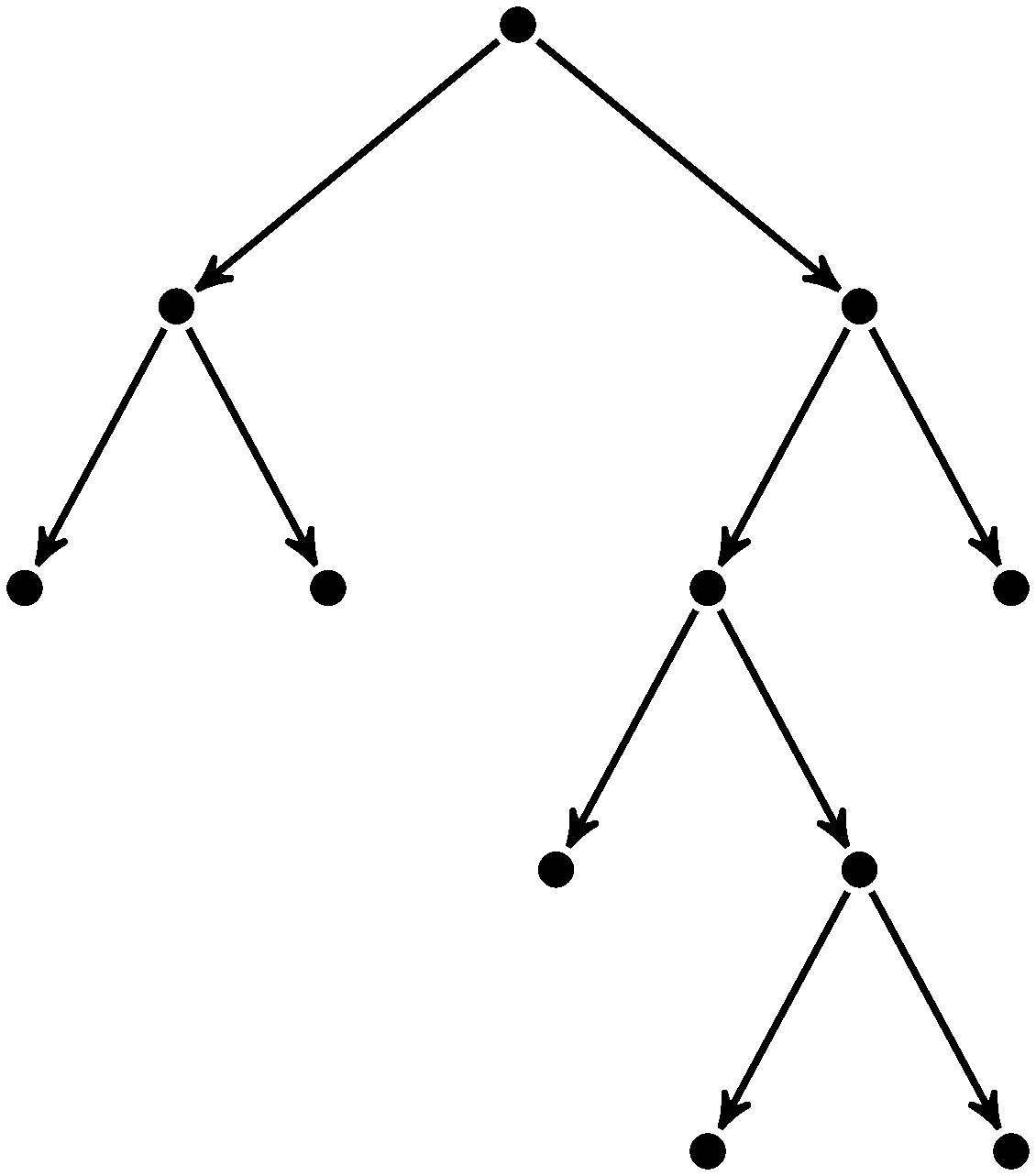}
\caption{Structure of an unlabeled binary tree}
\label{f10}
\end{figure}

Figure \ref{f10} shows the structure of a bare binary tree, and the intention is to find the corresponding natural number for a tree with that structure.

\begin{figure}
\centering
\subfloat[]{ 
\includegraphics{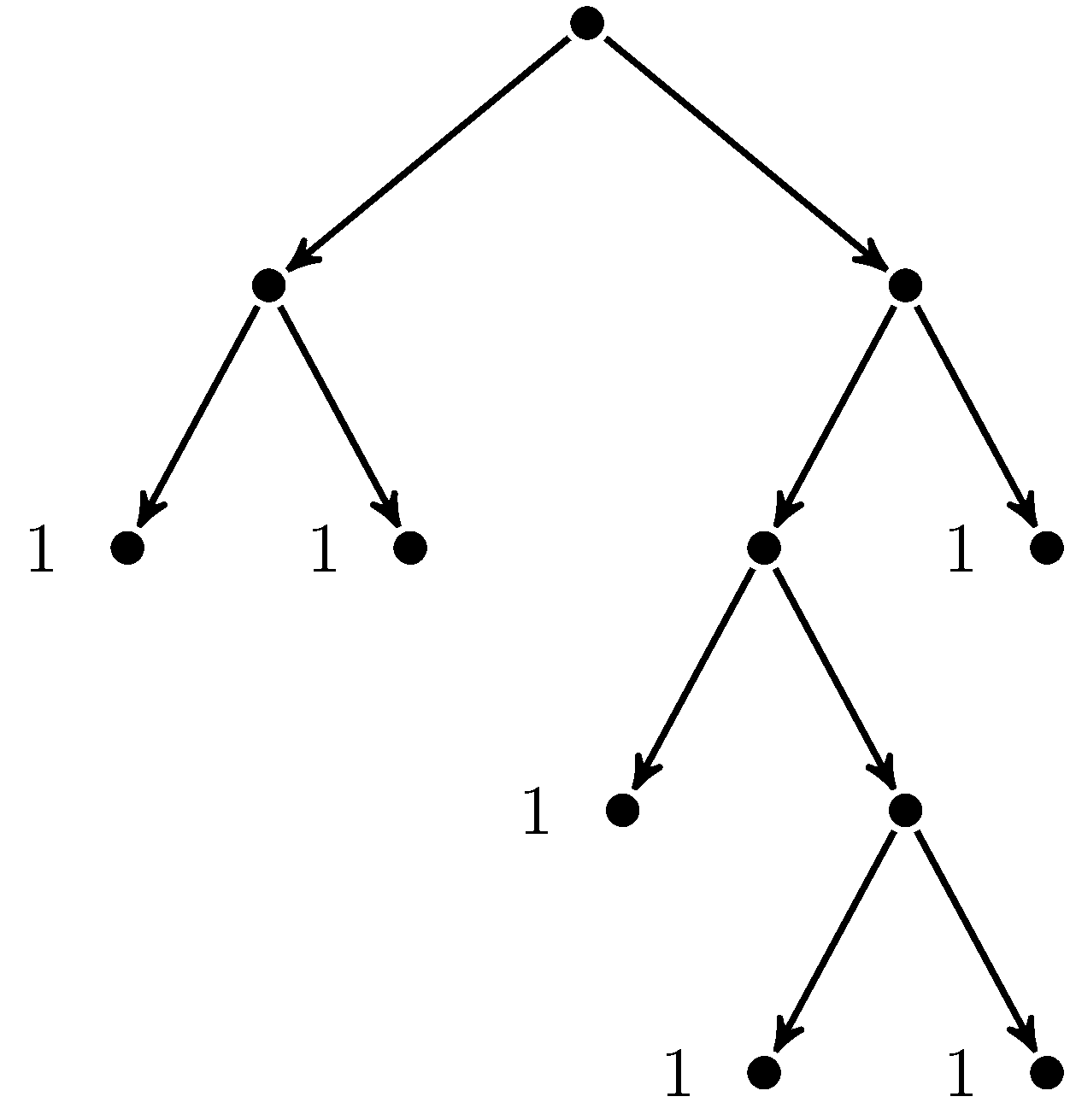}}\qquad
\subfloat[]{ 
\includegraphics{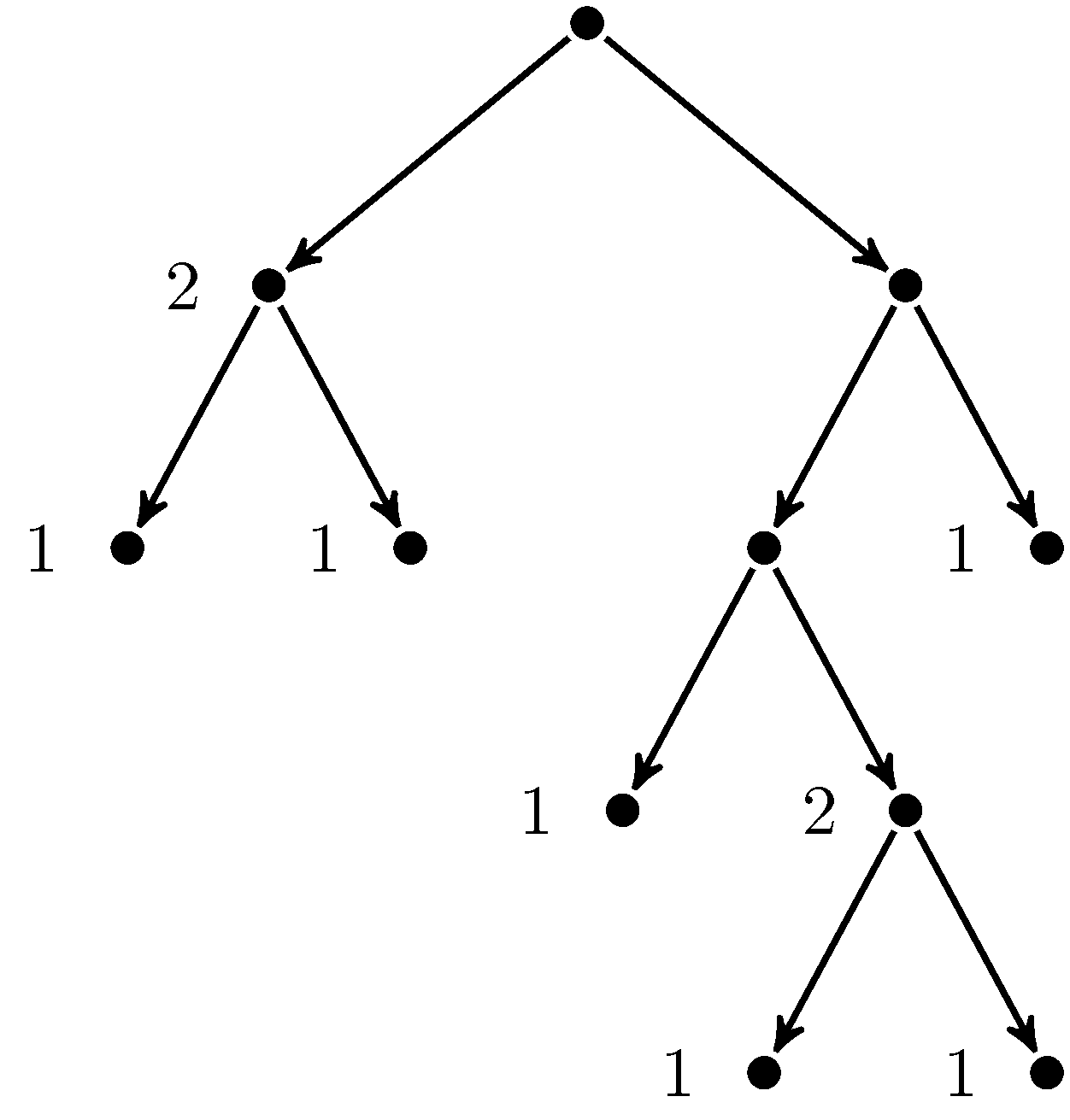}
} 
\subfloat[]{ 
\includegraphics{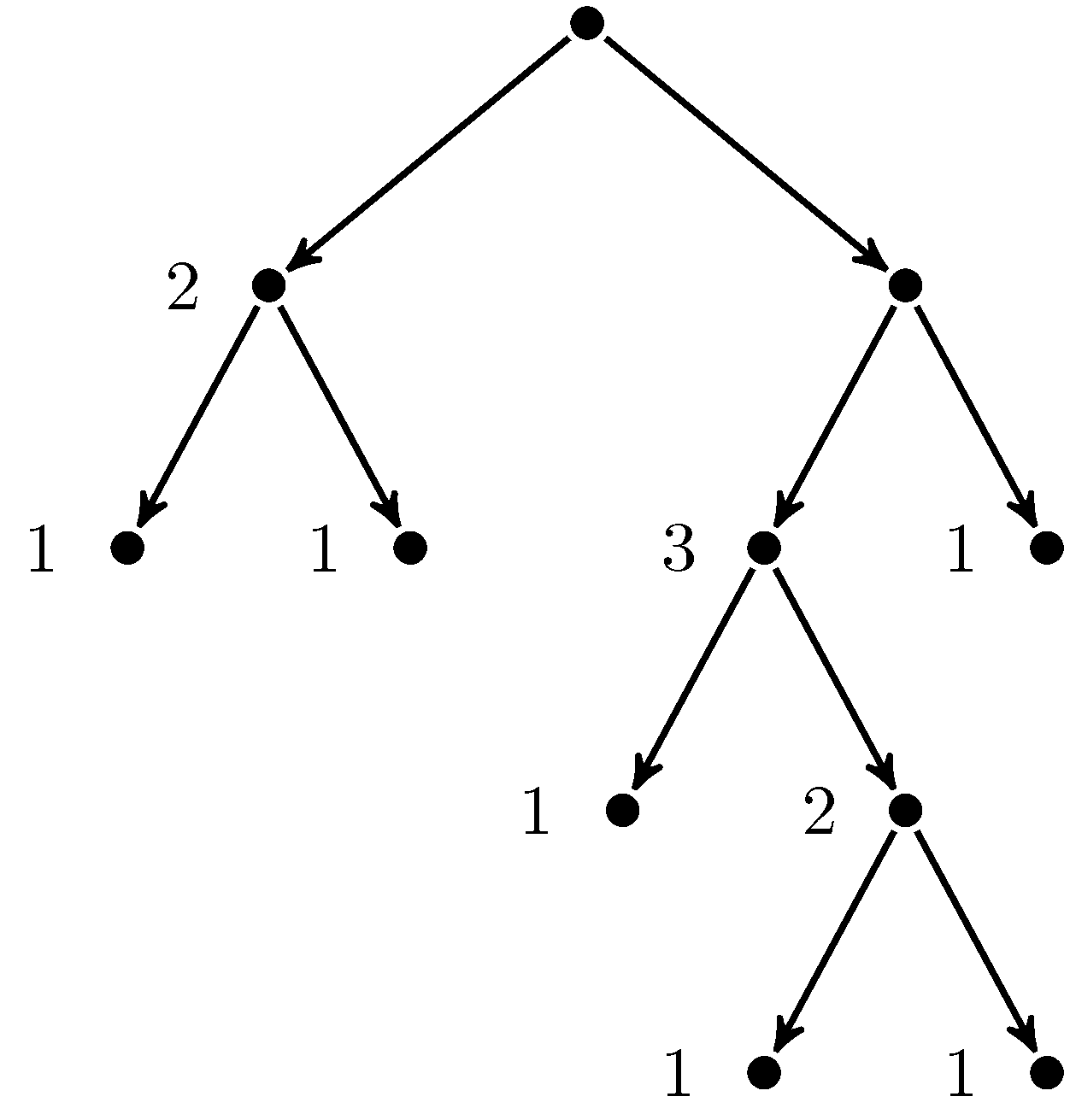} }\qquad
\subfloat[]{ 
\includegraphics{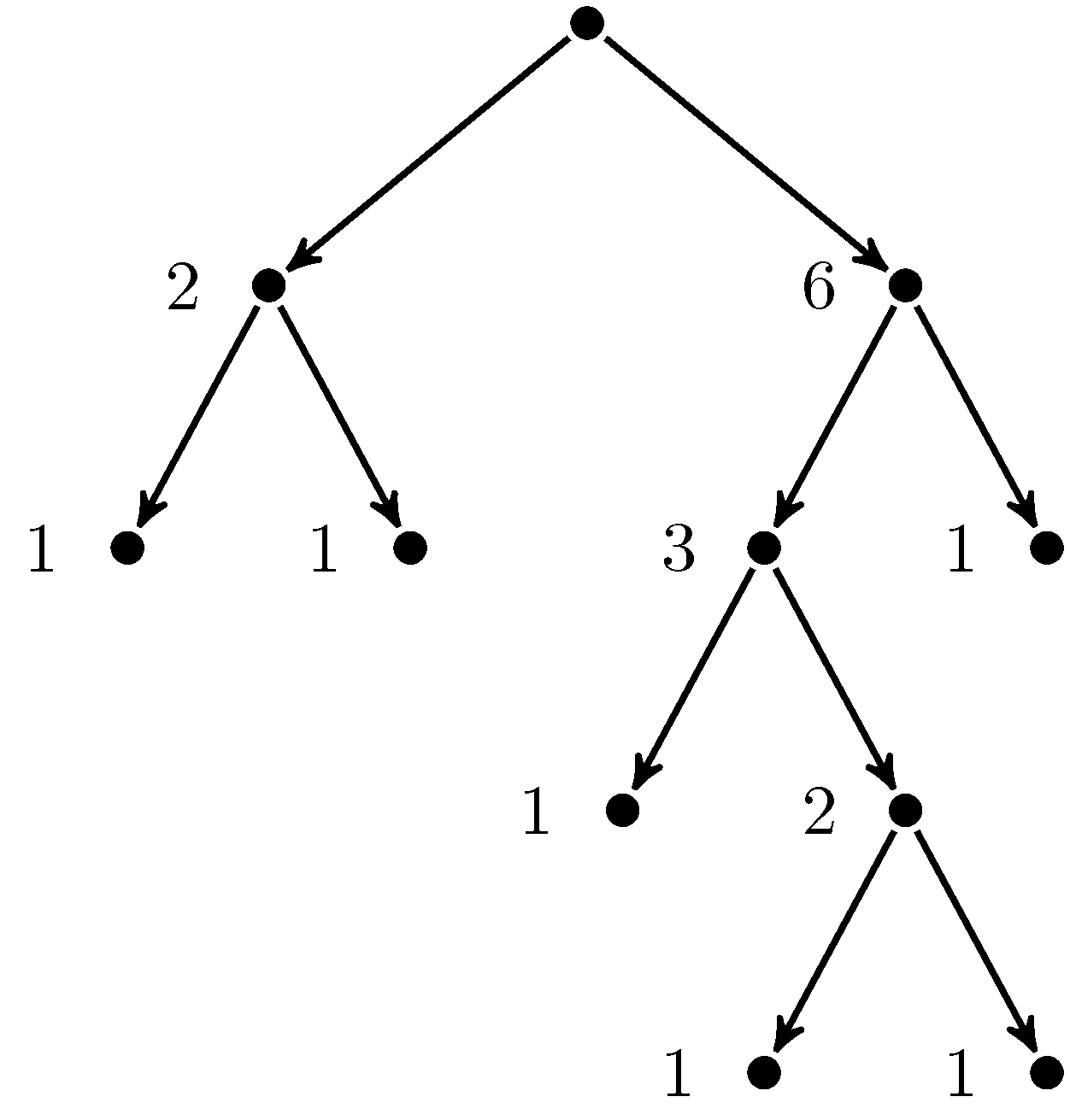} } \qquad
\subfloat[]{ 
\includegraphics{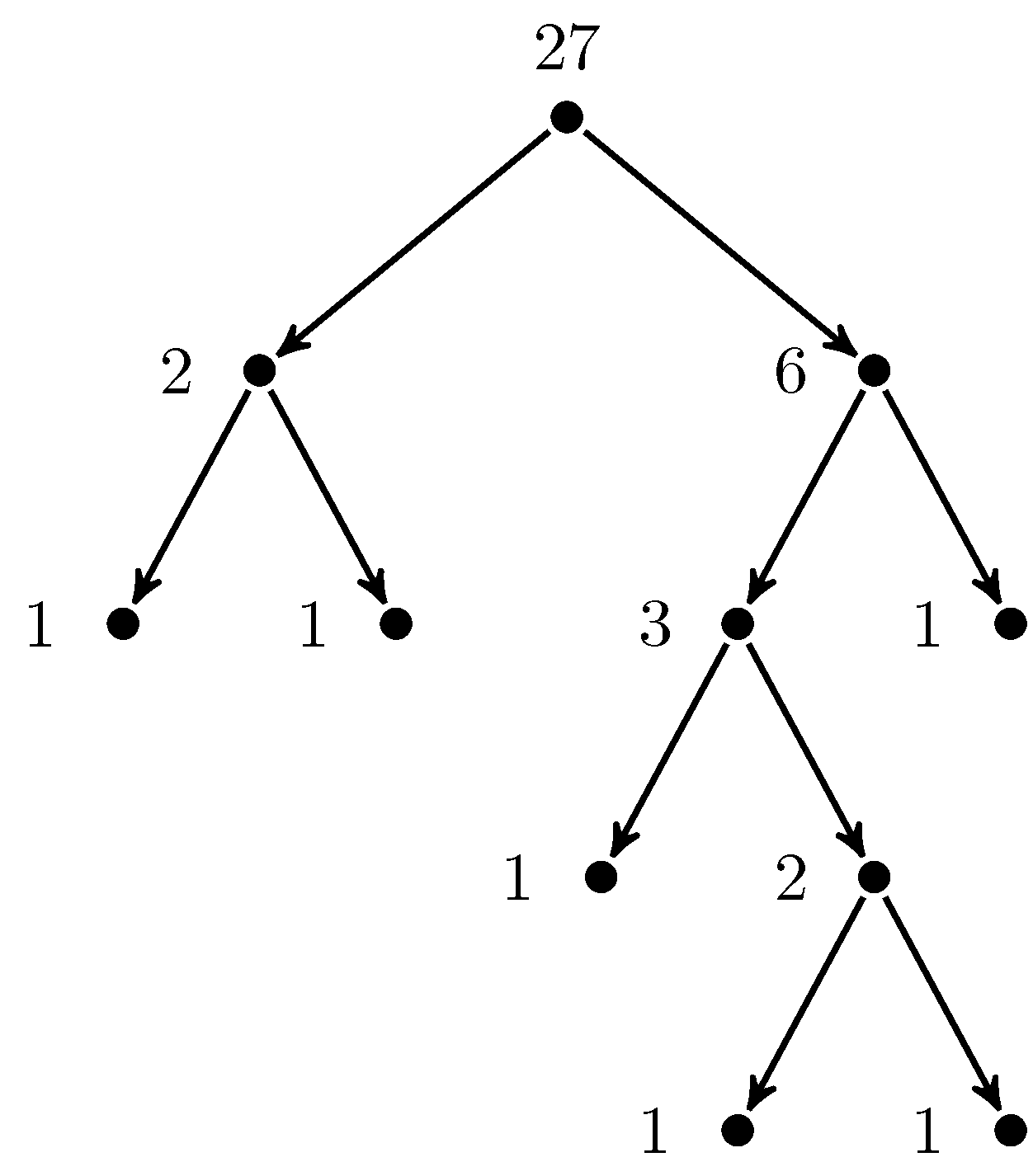}} 
\caption{Stages corresponding to the process applied to determine the number -- 27 -- corresponding to the binary tree whose structure was represented in figure \ref{f10}}
\label{f11}
\end{figure}

In subfigures 11(a), 11(b), 11(c), 11(d) and 11(e) of figure \ref{f11}, the successive stages presented show how it is possible to determine the natural number to which the bare binary tree in figure \ref{f10} corresponds.

\newpage
\vfill
\clearpage 

\begin{figure}[!h] 
\centering
\includegraphics{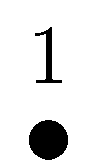}
\caption{Graphic representation of the binary tree corresponding to the natural number 1: a single node which is both a root node and a terminal node}
\label{f12}
\end{figure}

Clearly, the graphic representation of the binary tree corresponding to the natural number 1, according to the criterion used, consists of one dot for any node that is both a root and a terminal node. This is shown in figure \ref{f12}.\\

\subsection{Second Variant for the One-to-One Correspondence between Binary Trees and Natural Numbers}

One limitation or characteristic that might be considered undesirable of the one-to-one correspondence described in section 4.2 is that it necessarily requires distinguishing between left and right to specify the locations of the child nodes in the graphic representations. Binary trees are usually considered specific cases of graphs but in these mathematical entities this limitation does not exist.

This limitation can be prevented with the use of labeled binary trees. In this variant each node other than the root node can be labeled as \textbf{T} (for \textit{Type}) or \textbf{O} (for \textit{Order}) according to whether the natural number corresponding to the node (either type or order respectively) of the parent node of the node considered.

\begin{figure}[!h] 
\centering
\includegraphics{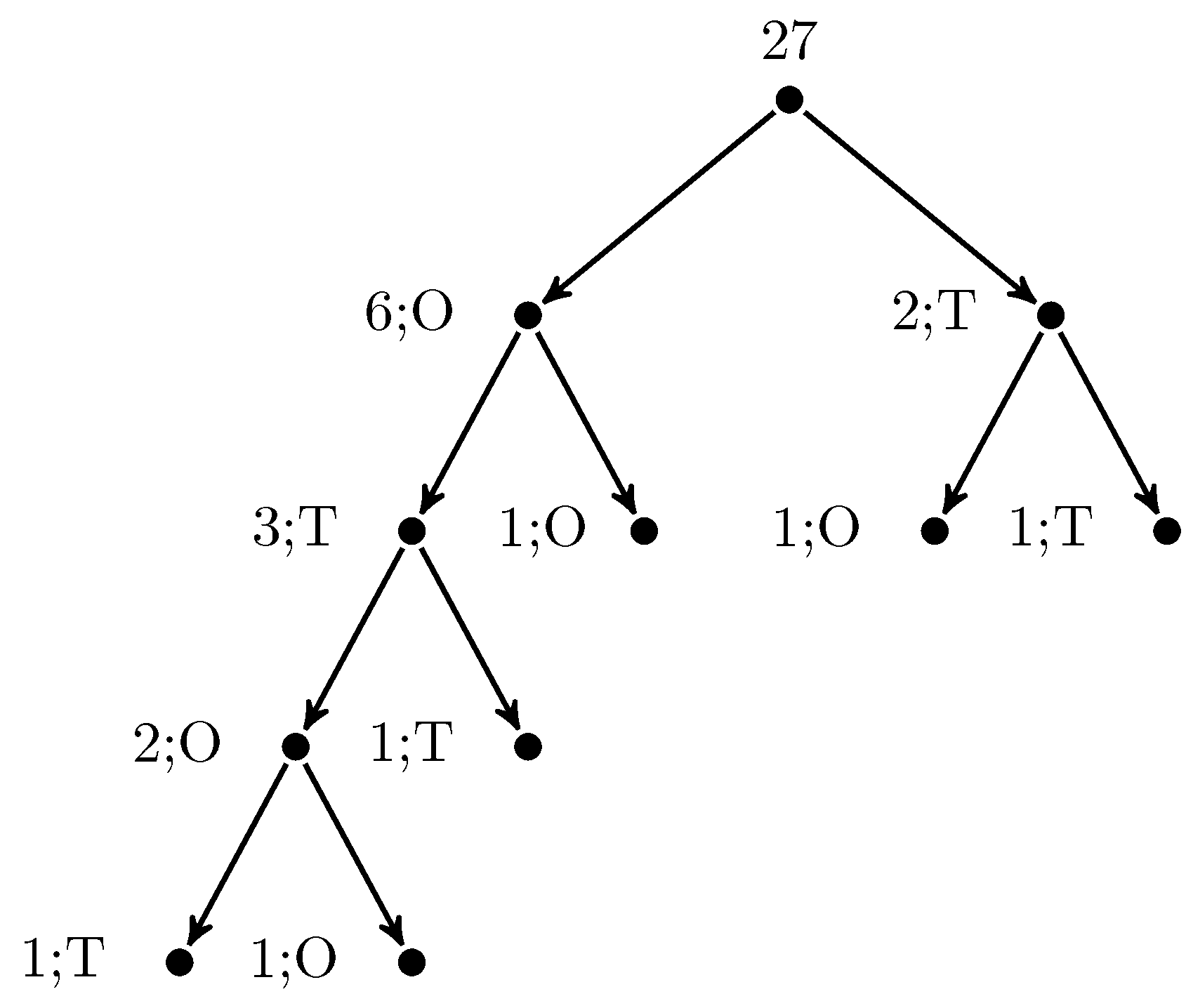}
\caption{Graphic representation of the binary tree corresponding to the natural number 27, according to the second variant described}
\label{f13}
\end{figure}

Figure \ref{f13} is a graphic representation, according to the second variant described here, of the binary tree for the number 27. Recall that figure \ref{f11} is the representation without the special tags.

It can observed in figure \ref{f13} that each child node is labeled \textbf{T} or \textbf{O} to specify whether the node corresponds to the type or to the order, respectively, of the natural number corresponding to the parent node. Therefore, it does not matter which node is placed to the left or to the right of the other. Figure \ref{f13} also indicates a) the number assigned to the root node (27), and b) the natural numbers corresponding to each child node. The latter are have been specified before the respective label for each node (T or O). 
\newpage 

 Of course, in this second variant an unnumbered binary tree with labels T and O can be also analyzed from the terminal nodes toward the root to determine the natural number corresponding to that tree.

\section{Prospects}

It should be observed that in some cases by using tables of different numerical sequences, such as those available in OEIS \cite{oeis1}, the use of the algorithmic approach specified in section 3 to determine the ordered pair corresponding to any natural number could be avoided. Suppose that one desires to find the ordered pair corresponding to the number $996$; that is, $(4, 275)$. A search could be carried out in sequences of natural numbers divisible by different numbers of prime numbers. If the number 996 is found in a sequence of natural numbers divisible by exactly three different primes, then one knows that the number is type 4, according to how the notion of type is characterized for natural numbers. If the order to which 996 corresponds in this sequence of natural numbers is not specified explicitly, it could be determined by counting. Nevertheless, that procedure can generate an undesirable dependence on the data available in the tables mentioned above.\\

Future articles will address the following topics:
\begin{enumerate}
\item a possible application of the results of this paper in the field of information compression
\item a possible use of binary trees to characterize elements belonging to sets of numbers other than that of natural numbers.
\end{enumerate}

\section{Appendix: List of the Ordered Pairs (Type, Order) Corresponding to the Natural Numbers up to 1080}   

\newpage
\noindent\begin{tabular}{cclcclcc}\cmidrule[\heavyrulewidth](lr){1-2}\cmidrule[\heavyrulewidth](lr){4-5}\cmidrule[\heavyrulewidth](lr){7-8}
Natural & (Type, Order) & &Natural & (Type, Order) & &Natural & (Type, Order) \\
number & & & number & & & number & \\\cmidrule[\lightrulewidth](lr){1-2}\cmidrule[\lightrulewidth](lr){4-5}\cmidrule[\lightrulewidth](lr){7-8}
1 &   --   & & 41 & (1,13) & & 81 & (2,10)\\
2 & (1,1) & & 42 & (4,2) & & 82 & (3,43)\\
3 & (1,2) & & 43 & (1,14) & & 83 & (1,23)\\
4 & (2,1) & & 44 & (3,20) & & 84 & (4,7)\\
5 & (1,3) & & 45 & (3,21) & & 85 & (3,44)\\
6 & (3,1) & & 46 & (3,22) & & 86 & (3,45)\\
7 & (1,4) & & 47 & (1,15) & & 87 & (3,46)\\
8 & (2,2) & & 48 & (3,23) & & 88 & (3,47)\\
9 & (2,3) & & 49 & (2,8) & & 89 & (1,24)\\
10 & (3,2) & & 50 & (3,24) & & 90 & (4,8)\\
11 & (1,5) & & 51 & (3,25) & & 91 & (3,48)\\
12 & (3,3) & & 52 & (3,26) & & 92 & (3,49)\\
13 & (1,6) & & 53 & (1,16) & & 93 & (3,50)\\
14 & (3,4) & & 54 & (3,27) & & 94 & (3,51)\\
15 & (3,5) & & 55 & (3,28) & & 95 & (3,52)\\
16 & (2,4) & & 56 & (3,29) & & 96 & (3,53)\\
17 & (1,7) & & 57 & (3,30) & & 97 & (1,25)\\
18 & (3,6) & & 58 & (3,31) & & 98 & (3,54)\\
19 & (1,8) & & 59 & (1,17) & & 99 & (3,55)\\
20 & (3,7) & & 60 & (4,3) & & 100 & (3,56)\\
21 & (3,8) & & 61 & (1,18) & & 101 & (1,26)\\
22 & (3,9) & & 62 & (3,32) & & 102 & (4,9)\\
23 & (1,9) & & 63 & (3,33) & & 103 & (1,27)\\
24 & (3,10) & & 64 & (2,9) & & 104 & (3,57)\\
25 & (2,5) & & 65 & (3,34) & & 105 & (4,10)\\
26 & (3,11) & & 66 & (4,4) & & 106 & (3,58)\\
27 & (2,6) & & 67 & (1,19) & & 107 & (1,28)\\
28 & (3,12) & & 68 & (3,35) & & 108 & (3,59)\\
29 & (1,10) & & 69 & (3,36) & & 109 & (1,29)\\
30 & (4,1) & & 70 & (4,5) & & 110 & (4,11)\\
31 & (1,11) & & 71 & (1,20) & & 111 & (3,60)\\
32 & (2,7) & & 72 & (3,37) & & 112 & (3,61)\\
33 & (3,13) & & 73 & (1,21) & & 113 & (1,30)\\
34 & (3,14) & & 74 & (3,38) & & 114 & (4,12)\\
35 & (3,15) & & 75 & (3,39) & & 115 & (3,62)\\
36 & (3,16) & & 76 & (3,40) & & 116 & (3,63)\\
37 & (1,12) & & 77 & (3,41) & & 117 & (3,64)\\
38 & (3,17) & & 78 & (4,6) & & 118 & (3,65)\\
39 & (3,18) & & 79 & (1,22) & & 119 & (3,66)\\
40 & (3,19) & & 80 & (3,42) & & 120 & (4,13)\\
\cmidrule[\heavyrulewidth](lr){1-2}\cmidrule[\heavyrulewidth](lr){4-5}\cmidrule[\heavyrulewidth](lr){7-8}
\end{tabular}
\newpage
\noindent\begin{tabular}{cclcclcc}\cmidrule[\heavyrulewidth](lr){1-2}\cmidrule[\heavyrulewidth](lr){4-5}\cmidrule[\heavyrulewidth](lr){7-8}
Natural & (Type, Order) & &Natural & (Type, Order) & &Natural & (Type, Order) \\
number & & & number & & & number & \\\cmidrule[\lightrulewidth](lr){1-2}\cmidrule[\lightrulewidth](lr){4-5}\cmidrule[\lightrulewidth](lr){7-8}
121 & (2,11) & & 161 & (3,89) & & 201 & (3,109)\\
122 & (3,67) & & 162 & (3,90) & & 202 & (3,110)\\
123 & (3,68) & & 163 & (1,38) & & 203 & (3,111)\\
124 & (3,69) & & 164 & (3,91) & & 204 & (4,32)\\
125 & (2,12) & & 165 & (4,22) & & 205 & (3,112)\\
126 & (4,14) & & 166 & (3,92) & & 206 & (3,113)\\
127 & (1,31) & & 167 & (1,39) & & 207 & (3,114)\\
128 & (2,13) & & 168 & (4,23) & & 208 & (3,115)\\
129 & (3,70) & & 169 & (2,14) & & 209 & (3,116)\\
130 & (4,15) & & 170 & (4,24) & & 210 & (5,1)\\
131 & (1,32) & & 171 & (3,93) & & 211 & (1,47)\\
132 & (4,16) & & 172 & (3,94) & & 212 & (3,117)\\
133 & (3,71) & & 173 & (1,40) & & 213 & (3,118)\\
134 & (3,72) & & 174 & (4,25) & & 214 & (3,119)\\
135 & (3,73) & & 175 & (3,95) & & 215 & (3,120)\\
136 & (3,74) & & 176 & (3,96) & & 216 & (3,121)\\
137 & (1,33) & & 177 & (3,97) & & 217 & (3,122)\\
138 & (4,17) & & 178 & (3,98) & & 218 & (3,123)\\
139 & (1,34) & & 179 & (1,41) & & 219 & (3,124)\\
140 & (4,18) & & 180 & (4,26) & & 220 & (4,33)\\
141 & (3,75) & & 181 & (1,42) & & 221 & (3,125)\\
142 & (3,76) & & 182 & (4,27) & & 222 & (4,34)\\
143 & (3,77) & & 183 & (3,99) & & 223 & (1,48)\\
144 & (3,78) & & 184 & (3,100) & & 224 & (3,126)\\
145 & (3,79) & & 185 & (3,101) & & 225 & (3,127)\\
146 & (3,80) & & 186 & (4,28) & & 226 & (3,128)\\
147 & (3,81) & & 187 & (3,102) & & 227 & (1,49)\\
148 & (3,82) & & 188 & (3,103) & & 228 & (4,35)\\
149 & (1,35) & & 189 & (3,104) & & 229 & (1,50)\\
150 & (4,19) & & 190 & (4,29) & & 230 & (4,36)\\
151 & (1,36) & & 191 & (1,43) & & 231 & (4,37)\\
152 & (3,83) & & 192 & (3,105) & & 232 & (3,129)\\
153 & (3,84) & & 193 & (1,44) & & 233 & (1,51)\\
154 & (4,20) & & 194 & (3,106) & & 234 & (4,38)\\
155 & (3,85) & & 195 & (4,30) & & 235 & (3,130)\\
156 & (4,21) & & 196 & (3,107) & & 236 & (3,131)\\
157 & (1,37) & & 197 & (1,45) & & 237 & (3,132)\\
158 & (3,86) & & 198 & (4,31) & & 238 & (4,39)\\
159 & (3,87) & & 199 & (1,46) & & 239 & (1,52)\\
160 & (3,88) & & 200 & (3,108) & & 240 & (4,40)\\
\cmidrule[\heavyrulewidth](lr){1-2}\cmidrule[\heavyrulewidth](lr){4-5}\cmidrule[\heavyrulewidth](lr){7-8}
\end{tabular}
\newpage
\noindent\begin{tabular}{cclcclcc}\cmidrule[\heavyrulewidth](lr){1-2}\cmidrule[\heavyrulewidth](lr){4-5}\cmidrule[\heavyrulewidth](lr){7-8}
Natural & (Type, Order) & &Natural & (Type, Order) & &Natural & (Type, Order) \\
number & & & number & & & number & \\\cmidrule[\lightrulewidth](lr){1-2}\cmidrule[\lightrulewidth](lr){4-5}\cmidrule[\lightrulewidth](lr){7-8}
241 & (1,53) & & 281 & (1,60) & & 321 & (3,173)\\
242 & (3,133) & & 282 & (4,52) & & 322 & (4,64)\\
243 & (2,15) & & 283 & (1,61) & & 323 & (3,174)\\
244 & (3,134) & & 284 & (3,153) & & 324 & (3,175)\\
245 & (3,135) & & 285 & (4,53) & & 325 & (3,176)\\
246 & (4,41) & & 286 & (4,54) & & 326 & (3,177)\\
247 & (3,136) & & 287 & (3,154) & & 327 & (3,178)\\
248 & (3,137) & & 288 & (3,155) & & 328 & (3,179)\\
249 & (3,138) & & 289 & (2,17) & & 329 & (3,180)\\
250 & (3,139) & & 290 & (4,55) & & 330 & (5,2)\\
251 & (1,54) & & 291 & (3,156) & & 331 & (1,67)\\
252 & (4,42) & & 292 & (3,157) & & 332 & (3,181)\\
253 & (3,140) & & 293 & (1,62) & & 333 & (3,182)\\
254 & (3,141) & & 294 & (4,56) & & 334 & (3,183)\\
255 & (4,43) & & 295 & (3,158) & & 335 & (3,184)\\
256 & (2,16) & & 296 & (3,159) & & 336 & (4,65)\\
257 & (1,55) & & 297 & (3,160) & & 337 & (1,68)\\
258 & (4,44) & & 298 & (3,161) & & 338 & (3,185)\\
259 & (3,142) & & 299 & (3,162) & & 339 & (3,186)\\
260 & (4,45) & & 300 & (4,57) & & 340 & (4,66)\\
261 & (3,143) & & 301 & (3,163) & & 341 & (3,187)\\
262 & (3,144) & & 302 & (3,164) & & 342 & (4,67)\\
263 & (1,56) & & 303 & (3,165) & & 343 & (2,18)\\
264 & (4,46) & & 304 & (3,166) & & 344 & (3,188)\\
265 & (3,145) & & 305 & (3,167) & & 345 & (4,68)\\
266 & (4,47) & & 306 & (4,58) & & 346 & (3,189)\\
267 & (3,146) & & 307 & (1,63) & & 347 & (1,69)\\
268 & (3,147) & & 308 & (4,59) & & 348 & (4,69)\\
269 & (1,57) & & 309 & (3,168) & & 349 & (1,70)\\
270 & (4,48) & & 310 & (4,60) & & 350 & (4,70)\\
271 & (1,58) & & 311 & (1,64) & & 351 & (3,190)\\
272 & (3,148) & & 312 & (4,61) & & 352 & (3,191)\\
273 & (4,49) & & 313 & (1,65) & & 353 & (1,71)\\
274 & (3,149) & & 314 & (3,169) & & 354 & (4,71)\\
275 & (3,150) & & 315 & (4,62) & & 355 & (3,192)\\
276 & (4,50) & & 316 & (3,170) & & 356 & (3,193)\\
277 & (1,59) & & 317 & (1,66) & & 357 & (4,72)\\
278 & (3,151) & & 318 & (4,63) & & 358 & (3,194)\\
279 & (3,152) & & 319 & (3,171) & & 359 & (1,72)\\
280 & (4,51) & & 320 & (3,172) & & 360 & (4,73)\\
\cmidrule[\heavyrulewidth](lr){1-2}\cmidrule[\heavyrulewidth](lr){4-5}\cmidrule[\heavyrulewidth](lr){7-8}
\end{tabular}
\newpage
\noindent\begin{tabular}{cclcclcc}\cmidrule[\heavyrulewidth](lr){1-2}\cmidrule[\heavyrulewidth](lr){4-5}\cmidrule[\heavyrulewidth](lr){7-8}
Natural & (Type, Order) & &Natural & (Type, Order) & &Natural & (Type, Order) \\
number & & & number & & & number & \\\cmidrule[\lightrulewidth](lr){1-2}\cmidrule[\lightrulewidth](lr){4-5}\cmidrule[\lightrulewidth](lr){7-8}
361 & (2,19) & & 401 & (1,79) & & 441 & (3,236)\\
362 & (3,195) & & 402 & (4,84) & & 442 & (4,97)\\
363 & (3,196) & & 403 & (3,217) & & 443 & (1,86)\\
364 & (4,74) & & 404 & (3,218) & & 444 & (4,98)\\
365 & (3,197) & & 405 & (3,219) & & 445 & (3,237)\\
366 & (4,75) & & 406 & (4,85) & & 446 & (3,238)\\
367 & (1,73) & & 407 & (3,220) & & 447 & (3,239)\\
368 & (3,198) & & 408 & (4,86) & & 448 & (3,240)\\
369 & (3,199) & & 409 & (1,80) & & 449 & (1,87)\\
370 & (4,76) & & 410 & (4,87) & & 450 & (4,99)\\
371 & (3,200) & & 411 & (3,221) & & 451 & (3,241)\\
372 & (4,77) & & 412 & (3,222) & & 452 & (3,242)\\
373 & (1,74) & & 413 & (3,223) & & 453 & (3,243)\\
374 & (4,78) & & 414 & (4,88) & & 454 & (3,244)\\
375 & (3,201) & & 415 & (3,224) & & 455 & (4,100)\\
376 & (3,202) & & 416 & (3,225) & & 456 & (4,101)\\
377 & (3,203) & & 417 & (3,226) & & 457 & (1,88)\\
378 & (4,79) & & 418 & (4,89) & & 458 & (3,245)\\
379 & (1,75) & & 419 & (1,81) & & 459 & (3,246)\\
380 & (4,80) & & 420 & (5,4) & & 460 & (4,102)\\
381 & (3,204) & & 421 & (1,82) & & 461 & (1,89)\\
382 & (3,205) & & 422 & (3,227) & & 462 & (5,5)\\
383 & (1,76) & & 423 & (3,228) & & 463 & (1,90)\\
384 & (3,206) & & 424 & (3,229) & & 464 & (3,247)\\
385 & (4,81) & & 425 & (3,230) & & 465 & (4,103)\\
386 & (3,207) & & 426 & (4,90) & & 466 & (3,248)\\
387 & (3,208) & & 427 & (3,231) & & 467 & (1,91)\\
388 & (3,209) & & 428 & (3,232) & & 468 & (4,104)\\
389 & (1,77) & & 429 & (4,91) & & 469 & (3,249)\\
390 & (5,3) & & 430 & (4,92) & & 470 & (4,105)\\
391 & (3,210) & & 431 & (1,83) & & 471 & (3,250)\\
392 & (3,211) & & 432 & (3,233) & & 472 & (3,251)\\
393 & (3,212) & & 433 & (1,84) & & 473 & (3,252)\\
394 & (3,213) & & 434 & (4,93) & & 474 & (4,106)\\
395 & (3,214) & & 435 & (4,94) & & 475 & (3,253)\\
396 & (4,82) & & 436 & (3,234) & & 476 & (4,107)\\
397 & (1,78) & & 437 & (3,235) & & 477 & (3,254)\\
398 & (3,215) & & 438 & (4,95) & & 478 & (3,255)\\
399 & (4,83) & & 439 & (1,85) & & 479 & (1,92)\\
400 & (3,216) & & 440 & (4,96) & & 480 & (4,108)\\
\cmidrule[\heavyrulewidth](lr){1-2}\cmidrule[\heavyrulewidth](lr){4-5}\cmidrule[\heavyrulewidth](lr){7-8}
\end{tabular}
\newpage
\noindent\begin{tabular}{cclcclcc}\cmidrule[\heavyrulewidth](lr){1-2}\cmidrule[\heavyrulewidth](lr){4-5}\cmidrule[\heavyrulewidth](lr){7-8}
Natural & (Type, Order) & &Natural & (Type, Order) & &Natural & (Type, Order) \\
number & & & number & & & number & \\\cmidrule[\lightrulewidth](lr){1-2}\cmidrule[\lightrulewidth](lr){4-5}\cmidrule[\lightrulewidth](lr){7-8}
481 & (3,256) & & 521 & (1,98) & & 561 & (4,132)\\
482 & (3,257) & & 522 & (4,120) & & 562 & (3,299)\\
483 & (4,109) & & 523 & (1,99) & & 563 & (1,103)\\
484 & (3,258) & & 524 & (3,278) & & 564 & (4,133)\\
485 & (3,259) & & 525 & (4,121) & & 565 & (3,300)\\
486 & (3,260) & & 526 & (3,279) & & 566 & (3,301)\\
487 & (1,93) & & 527 & (3,280) & & 567 & (3,302)\\
488 & (3,261) & & 528 & (4,122) & & 568 & (3,303)\\
489 & (3,262) & & 529 & (2,21) & & 569 & (1,104)\\
490 & (4,110) & & 530 & (4,123) & & 570 & (5,8)\\
491 & (1,94) & & 531 & (3,281) & & 571 & (1,105)\\
492 & (4,111) & & 532 & (4,124) & & 572 & (4,134)\\
493 & (3,263) & & 533 & (3,282) & & 573 & (3,304)\\
494 & (4,112) & & 534 & (4,125) & & 574 & (4,135)\\
495 & (4,113) & & 535 & (3,283) & & 575 & (3,305)\\
496 & (3,264) & & 536 & (3,284) & & 576 & (3,306)\\
497 & (3,265) & & 537 & (3,285) & & 577 & (1,106)\\
498 & (4,114) & & 538 & (3,286) & & 578 & (3,307)\\
499 & (1,95) & & 539 & (3,287) & & 579 & (3,308)\\
500 & (3,266) & & 540 & (4,126) & & 580 & (4,136)\\
501 & (3,267) & & 541 & (1,100) & & 581 & (3,309)\\
502 & (3,268) & & 542 & (3,288) & & 582 & (4,137)\\
503 & (1,96) & & 543 & (3,289) & & 583 & (3,310)\\
504 & (4,115) & & 544 & (3,290) & & 584 & (3,311)\\
505 & (3,269) & & 545 & (3,291) & & 585 & (4,138)\\
506 & (4,116) & & 546 & (5,7) & & 586 & (3,312)\\
507 & (3,270) & & 547 & (1,101) & & 587 & (1,107)\\
508 & (3,271) & & 548 & (3,292) & & 588 & (4,139)\\
509 & (1,97) & & 549 & (3,293) & & 589 & (3,313)\\
510 & (5,6) & & 550 & (4,127) & & 590 & (4,140)\\
511 & (3,272) & & 551 & (3,294) & & 591 & (3,314)\\
512 & (2,20) & & 552 & (4,128) & & 592 & (3,315)\\
513 & (3,273) & & 553 & (3,295) & & 593 & (1,108)\\
514 & (3,274) & & 554 & (3,296) & & 594 & (4,141)\\
515 & (3,275) & & 555 & (4,129) & & 595 & (4,142)\\
516 & (4,117) & & 556 & (3,297) & & 596 & (3,316)\\
517 & (3,276) & & 557 & (1,102) & & 597 & (3,317)\\
518 & (4,118) & & 558 & (4,130) & & 598 & (4,143)\\
519 & (3,277) & & 559 & (3,298) & & 599 & (1,109)\\
520 & (4,119) & & 560 & (4,131) & & 600 & (4,144)\\
\cmidrule[\heavyrulewidth](lr){1-2}\cmidrule[\heavyrulewidth](lr){4-5}\cmidrule[\heavyrulewidth](lr){7-8}
\end{tabular}
\newpage
\noindent\begin{tabular}{cclcclcc}\cmidrule[\heavyrulewidth](lr){1-2}\cmidrule[\heavyrulewidth](lr){4-5}\cmidrule[\heavyrulewidth](lr){7-8}
Natural & (Type, Order) & &Natural & (Type, Order) & &Natural & (Type, Order) \\
number & & & number & & & number & \\\cmidrule[\lightrulewidth](lr){1-2}\cmidrule[\lightrulewidth](lr){4-5}\cmidrule[\lightrulewidth](lr){7-8}
601 & (1,110) & & 641 & (1,116) & & 681 & (3,353)\\
602 & (4,145) & & 642 & (4,158) & & 682 & (4,173)\\
603 & (3,318) & & 643 & (1,117) & & 683 & (1,124)\\
604 & (3,319) & & 644 & (4,159) & & 684 & (4,174)\\
605 & (3,320) & & 645 & (4,160) & & 685 & (3,354)\\
606 & (4,146) & & 646 & (4,161) & & 686 & (3,355)\\
607 & (1,111) & & 647 & (1,118) & & 687 & (3,356)\\
608 & (3,321) & & 648 & (3,337) & & 688 & (3,357)\\
609 & (4,147) & & 649 & (3,338) & & 689 & (3,358)\\
610 & (4,148) & & 650 & (4,162) & & 690 & (5,11)\\
611 & (3,322) & & 651 & (4,163) & & 691 & (1,125)\\
612 & (4,149) & & 652 & (3,339) & & 692 & (3,359)\\
613 & (1,112) & & 653 & (1,119) & & 693 & (4,175)\\
614 & (3,323) & & 654 & (4,164) & & 694 & (3,360)\\
615 & (4,150) & & 655 & (3,340) & & 695 & (3,361)\\
616 & (4,151) & & 656 & (3,341) & & 696 & (4,176)\\
617 & (1,113) & & 657 & (3,342) & & 697 & (3,362)\\
618 & (4,152) & & 658 & (4,165) & & 698 & (3,363)\\
619 & (1,114) & & 659 & (1,120) & & 699 & (3,364)\\
620 & (4,153) & & 660 & (5,10) & & 700 & (4,177)\\
621 & (3,324) & & 661 & (1,121) & & 701 & (1,126)\\
622 & (3,325) & & 662 & (3,343) & & 702 & (4,178)\\
623 & (3,326) & & 663 & (4,166) & & 703 & (3,365)\\
624 & (4,154) & & 664 & (3,344) & & 704 & (3,366)\\
625 & (2,22) & & 665 & (4,167) & & 705 & (4,179)\\
626 & (3,327) & & 666 & (4,168) & & 706 & (3,367)\\
627 & (4,155) & & 667 & (3,345) & & 707 & (3,368)\\
628 & (3,328) & & 668 & (3,346) & & 708 & (4,180)\\
629 & (3,329) & & 669 & (3,347) & & 709 & (1,127)\\
630 & (5,9) & & 670 & (4,169) & & 710 & (4,181)\\
631 & (1,115) & & 671 & (3,348) & & 711 & (3,369)\\
632 & (3,330) & & 672 & (4,170) & & 712 & (3,370)\\
633 & (3,331) & & 673 & (1,122) & & 713 & (3,371)\\
634 & (3,332) & & 674 & (3,349) & & 714 & (5,12)\\
635 & (3,333) & & 675 & (3,350) & & 715 & (4,182)\\
636 & (4,156) & & 676 & (3,351) & & 716 & (3,372)\\
637 & (3,334) & & 677 & (1,123) & & 717 & (3,373)\\
638 & (4,157) & & 678 & (4,171) & & 718 & (3,374)\\
639 & (3,335) & & 679 & (3,352) & & 719 & (1,128)\\
640 & (3,336) & & 680 & (4,172) & & 720 & (4,183)\\
\cmidrule[\heavyrulewidth](lr){1-2}\cmidrule[\heavyrulewidth](lr){4-5}\cmidrule[\heavyrulewidth](lr){7-8}
\end{tabular}
\newpage
\noindent\begin{tabular}{cclcclcc}\cmidrule[\heavyrulewidth](lr){1-2}\cmidrule[\heavyrulewidth](lr){4-5}\cmidrule[\heavyrulewidth](lr){7-8}
Natural & (Type, Order) & &Natural & (Type, Order) & &Natural & (Type, Order) \\
number & & & number & & & number & \\\cmidrule[\lightrulewidth](lr){1-2}\cmidrule[\lightrulewidth](lr){4-5}\cmidrule[\lightrulewidth](lr){7-8}
721 & (3,375) & & 761 & (1,135) & & 801 & (3,415)\\
722 & (3,376) & & 762 & (4,200) & & 802 & (3,416)\\
723 & (3,377) & & 763 & (3,392) & & 803 & (3,417)\\
724 & (3,378) & & 764 & (3,393) & & 804 & (4,209)\\
725 & (3,379) & & 765 & (4,201) & & 805 & (4,210)\\
726 & (4,184) & & 766 & (3,394) & & 806 & (4,211)\\
727 & (1,129) & & 767 & (3,395) & & 807 & (3,418)\\
728 & (4,185) & & 768 & (3,396) & & 808 & (3,419)\\
729 & (2,23) & & 769 & (1,136) & & 809 & (1,140)\\
730 & (4,186) & & 770 & (5,13) & & 810 & (4,212)\\
731 & (3,380) & & 771 & (3,397) & & 811 & (1,141)\\
732 & (4,187) & & 772 & (3,398) & & 812 & (4,213)\\
733 & (1,130) & & 773 & (1,137) & & 813 & (3,420)\\
734 & (3,381) & & 774 & (4,202) & & 814 & (4,214)\\
735 & (4,188) & & 775 & (3,399) & & 815 & (3,421)\\
736 & (3,382) & & 776 & (3,400) & & 816 & (4,215)\\
737 & (3,383) & & 777 & (4,203) & & 817 & (3,422)\\
738 & (4,189) & & 778 & (3,401) & & 818 & (3,423)\\
739 & (1,131) & & 779 & (3,402) & & 819 & (4,216)\\
740 & (4,190) & & 780 & (5,14) & & 820 & (4,217)\\
741 & (4,191) & & 781 & (3,403) & & 821 & (1,142)\\
742 & (4,192) & & 782 & (4,204) & & 822 & (4,218)\\
743 & (1,132) & & 783 & (3,404) & & 823 & (1,143)\\
744 & (4,193) & & 784 & (3,405) & & 824 & (3,424)\\
745 & (3,384) & & 785 & (3,406) & & 825 & (4,219)\\
746 & (3,385) & & 786 & (4,205) & & 826 & (4,220)\\
747 & (3,386) & & 787 & (1,138) & & 827 & (1,144)\\
748 & (4,194) & & 788 & (3,407) & & 828 & (4,221)\\
749 & (3,387) & & 789 & (3,408) & & 829 & (1,145)\\
750 & (4,195) & & 790 & (4,206) & & 830 & (4,222)\\
751 & (1,133) & & 791 & (3,409) & & 831 & (3,425)\\
752 & (3,388) & & 792 & (4,207) & & 832 & (3,426)\\
753 & (3,389) & & 793 & (3,410) & & 833 & (3,427)\\
754 & (4,196) & & 794 & (3,411) & & 834 & (4,223)\\
755 & (3,390) & & 795 & (4,208) & & 835 & (3,428)\\
756 & (4,197) & & 796 & (3,412) & & 836 & (4,224)\\
757 & (1,134) & & 797 & (1,139) & & 837 & (3,429)\\
758 & (3,391) & & 798 & (5,15) & & 838 & (3,430)\\
759 & (4,198) & & 799 & (3,413) & & 839 & (1,146)\\
760 & (4,199) & & 800 & (3,414) & & 840 & (5,16)\\
\cmidrule[\heavyrulewidth](lr){1-2}\cmidrule[\heavyrulewidth](lr){4-5}\cmidrule[\heavyrulewidth](lr){7-8}
\end{tabular}
\newpage
\noindent\begin{tabular}{cclcclcc}\cmidrule[\heavyrulewidth](lr){1-2}\cmidrule[\heavyrulewidth](lr){4-5}\cmidrule[\heavyrulewidth](lr){7-8}
Natural & (Type, Order) & &Natural & (Type, Order) & &Natural & (Type, Order) \\
number & & & number & & & number & \\\cmidrule[\lightrulewidth](lr){1-2}\cmidrule[\lightrulewidth](lr){4-5}\cmidrule[\lightrulewidth](lr){7-8}
841 & (2,24) & & 881 & (1,152) & & 921 & (3,470)\\
842 & (3,431) & & 882 & (4,236) & & 922 & (3,471)\\
843 & (3,432) & & 883 & (1,153) & & 923 & (3,472)\\
844 & (3,433) & & 884 & (4,237) & & 924 & (5,20)\\
845 & (3,434) & & 885 & (4,238) & & 925 & (3,473)\\
846 & (4,225) & & 886 & (3,452) & & 926 & (3,474)\\
847 & (3,435) & & 887 & (1,154) & & 927 & (3,475)\\
848 & (3,436) & & 888 & (4,239) & & 928 & (3,476)\\
849 & (3,437) & & 889 & (3,453) & & 929 & (1,158)\\
850 & (4,226) & & 890 & (4,240) & & 930 & (5,21)\\
851 & (3,438) & & 891 & (3,454) & & 931 & (3,477)\\
852 & (4,227) & & 892 & (3,455) & & 932 & (3,478)\\
853 & (1,147) & & 893 & (3,456) & & 933 & (3,479)\\
854 & (4,228) & & 894 & (4,241) & & 934 & (3,480)\\
855 & (4,229) & & 895 & (3,457) & & 935 & (4,251)\\
856 & (3,439) & & 896 & (3,458) & & 936 & (4,252)\\
857 & (1,148) & & 897 & (4,242) & & 937 & (1,159)\\
858 & (5,17) & & 898 & (3,459) & & 938 & (4,253)\\
859 & (1,149) & & 899 & (3,460) & & 939 & (3,481)\\
860 & (4,230) & & 900 & (4,243) & & 940 & (4,254)\\
861 & (4,231) & & 901 & (3,461) & & 941 & (1,160)\\
862 & (3,440) & & 902 & (4,244) & & 942 & (4,255)\\
863 & (1,150) & & 903 & (4,245) & & 943 & (3,482)\\
864 & (3,441) & & 904 & (3,462) & & 944 & (3,483)\\
865 & (3,442) & & 905 & (3,463) & & 945 & (4,256)\\
866 & (3,443) & & 906 & (4,246) & & 946 & (4,257)\\
867 & (3,444) & & 907 & (1,155) & & 947 & (1,161)\\
868 & (4,232) & & 908 & (3,464) & & 948 & (4,258)\\
869 & (3,445) & & 909 & (3,465) & & 949 & (3,484)\\
870 & (5,18) & & 910 & (5,19) & & 950 & (4,259)\\
871 & (3,446) & & 911 & (1,156) & & 951 & (3,485)\\
872 & (3,447) & & 912 & (4,247) & & 952 & (4,260)\\
873 & (3,448) & & 913 & (3,466) & & 953 & (1,162)\\
874 & (4,233) & & 914 & (3,467) & & 954 & (4,261)\\
875 & (3,449) & & 915 & (4,248) & & 955 & (3,486)\\
876 & (4,234) & & 916 & (3,468) & & 956 & (3,487)\\
877 & (1,151) & & 917 & (3,469) & & 957 & (4,262)\\
878 & (3,450) & & 918 & (4,249) & & 958 & (3,488)\\
879 & (3,451) & & 919 & (1,157) & & 959 & (3,489)\\
880 & (4,235) & & 920 & (4,250) & & 960 & (4,263)\\
\cmidrule[\heavyrulewidth](lr){1-2}\cmidrule[\heavyrulewidth](lr){4-5}\cmidrule[\heavyrulewidth](lr){7-8}
\end{tabular}
\newpage
\noindent\begin{tabular}{cclcclcc}\cmidrule[\heavyrulewidth](lr){1-2}\cmidrule[\heavyrulewidth](lr){4-5}\cmidrule[\heavyrulewidth](lr){7-8}
Natural & (Type, Order) & &Natural & (Type, Order) & &Natural & (Type, Order) \\
number & & & number & & & number & \\\cmidrule[\lightrulewidth](lr){1-2}\cmidrule[\lightrulewidth](lr){4-5}\cmidrule[\lightrulewidth](lr){7-8}
961 & (2,25) & & 1001 & (4,276) & & 1041 & (3,522)\\
962 & (4,264) & & 1002 & (4,277) & & 1042 & (3,523)\\
963 & (3,490) & & 1003 & (3,509) & & 1043 & (3,524)\\
964 & (3,491) & & 1004 & (3,510) & & 1044 & (4,294)\\
965 & (3,492) & & 1005 & (4,278) & & 1045 & (4,295)\\
966 & (5,22) & & 1006 & (3,511) & & 1046 & (3,525)\\
967 & (1,163) & & 1007 & (3,512) & & 1047 & (3,526)\\
968 & (3,493) & & 1008 & (4,279) & & 1048 & (3,527)\\
969 & (4,265) & & 1009 & (1,169) & & 1049 & (1,176)\\
970 & (4,266) & & 1010 & (4,280) & & 1050 & (5,25)\\
971 & (1,164) & & 1011 & (3,513) & & 1051 & (1,177)\\
972 & (3,494) & & 1012 & (4,281) & & 1052 & (3,528)\\
973 & (3,495) & & 1013 & (1,170) & & 1053 & (3,529)\\
974 & (3,496) & & 1014 & (4,282) & & 1054 & (4,296)\\
975 & (4,267) & & 1015 & (4,283) & & 1055 & (3,530)\\
976 & (3,497) & & 1016 & (3,514) & & 1056 & (4,297)\\
977 & (1,165) & & 1017 & (3,515) & & 1057 & (3,531)\\
978 & (4,268) & & 1018 & (3,516) & & 1058 & (3,532)\\
979 & (3,498) & & 1019 & (1,171) & & 1059 & (3,533)\\
980 & (4,269) & & 1020 & (5,24) & & 1060 & (4,298)\\
981 & (3,499) & & 1021 & (1,172) & & 1061 & (1,178)\\
982 & (3,500) & & 1022 & (4,284) & & 1062 & (4,299)\\
983 & (1,166) & & 1023 & (4,285) & & 1063 & (1,179)\\
984 & (4,270) & & 1024 & (2,26) & & 1064 & (4,300)\\
985 & (3,501) & & 1025 & (3,517) & & 1065 & (4,301)\\
986 & (4,271) & & 1026 & (4,286) & & 1066 & (4,302)\\
987 & (4,272) & & 1027 & (3,518) & & 1067 & (3,534)\\
988 & (4,273) & & 1028 & (3,519) & & 1068 & (4,303)\\
989 & (3,502) & & 1029 & (3,520) & & 1069 & (1,180)\\
990 & (5,23) & & 1030 & (4,287) & & 1070 & (4,304)\\
991 & (1,167) & & 1031 & (1,173) & & 1071 & (4,305)\\
992 & (3,503) & & 1032 & (4,288) & & 1072 & (3,535)\\
993 & (3,504) & & 1033 & (1,174) & & 1073 & (3,536)\\
994 & (4,274) & & 1034 & (4,289) & & 1074 & (4,306)\\
995 & (3,505) & & 1035 & (4,290) & & 1075 & (3,537)\\
996 & (4,275) & & 1036 & (4,291) & & 1076 & (3,538)\\
997 & (1,168) & & 1037 & (3,521) & & 1077 & (3,539)\\
998 & (3,506) & & 1038 & (4,292) & & 1078 & (4,307)\\
999 & (3,507) & & 1039 & (1,175) & & 1079 & (3,540)\\
1000 & (3,508) & & 1040 & (4,293) & & 1080 & (4,308)\\
\cmidrule[\heavyrulewidth](lr){1-2}\cmidrule[\heavyrulewidth](lr){4-5}\cmidrule[\heavyrulewidth](lr){7-8}
\end{tabular}

\begin{thebibliography}{99}
 \bibitem[1]{tree1}D. Knuth, The Art of Computer Programming, Vol. 1, Third Edition. Pearson Education, New Jersey, 1997.
 \bibitem[2]{tree2}R. Garnier and J. Taylor, Discrete Mathematics: Proofs, Structures and Applications, Third Edition. CRC Press, Boca Raton, 2009.
 \bibitem[3]{tree3}D. R. Mazur, Combinatorics: A Guided Tour. Mathematical Association of America, Washington D. C., 2010.
 \bibitem[4]{tree4}K. Rosen, Discrete Mathematics and Its Applications, Seventh Edition. McGraw-Hill Science, New York, 2011.
 \bibitem[5]{app1} S. Russell and P. Norvig, Artificial Intelligence: A Modern Approach, Third (International) Edition. Pearson Education, Essex, 2010. 
 \bibitem[6]{app2}D. Knuth, The Art Of Computer Programming, Vol. 3, Third Edition. Pearson Education, New Jersey, 2014.
 \bibitem[7]{app3}L. Rokach and O. Maimon, Data Mining with Decision Trees: Theory and Applications. World Scientific Publication Company, New Jersey, 2008.
 \bibitem[8]{app4}J. W. J. Williams, \emph{Algorithm 232 - Heapsort}, Communications of the ACM \textbf{7} (6), 347--348, 1964.
 \bibitem[9]{app5}D. Huffman, \emph{A Method for the Construction of Minimum-Redundancy Codes}, Proceedings of the IRE \textbf{40} (9), 1098--1101, 1952.
 \bibitem[10]{app6}C. Bishop, Pattern Recognition and Machine Learning. Springer Science + Business, New York, 2004.
 \bibitem[11]{app7}R. O. Duda, P. E. Hart and D. G. Stork, Pattern Classification. John Wiley \& Sons, New York, 2001.
 \bibitem[12]{prev1}D. W. Matula, \emph{A Natural Rooted Tree Enumeration by Prime Factorization}, SIAM Rev. \textbf{10}, 1968.
 \bibitem[13]{prev2} D. W. Matula and Z. Chen, \emph{Precise and Concise Graphical Representation of the Natural Numbers}, 26th IEEE Symposium on Computer Arithmetic (ARITH), IEEE Xplore \textbf{21}, 100--103, October 2019. \bibitem[14]{prev3} I. Gutman and A. Ivíc, \emph{On Matula Numbers}, Discrete Mathematics \textbf{150.1} (1996): 131--142, 1996.
 \bibitem[15]{prev4} A. Burgos, \emph{On Natural Numbers as Trees: How Many Primes Are There?}. Japan Conference on Graph Theory and Combinatorics, May 2014.
 \bibitem[16]{prev5} P. R. Cappello, \emph{A New Bijection between Natural Numbers and Rooted Trees}, 4th SIAM Conf. on Discrete Mathematics, 1988.
 \bibitem[17]{prev6} F. Göbel, \emph{On a 1-1 Correspondence between Rooted Trees and Natural Numbers}, Journal of Combinatorial Theory, Series B \textbf{29.1} (1980): 141--143, 1980.
 \bibitem[18]{prev7} E. Deutsch. \emph{Rooted Tree Statistics from Matula Numbers}, Discrete Applied Mathematics \textbf{160.15} (2012): 2314--2322, 2012.
 \bibitem[19]{oeis1} OEIS Foundation Inc. \emph{The On-Line Encyclopedia of Integer Sequences}, https://oeis.org (2020).
 
\end{thebibliography}
\end{document}